\documentclass{ieeeaccess}
\usepackage{cite}
\usepackage{amsmath,amssymb,amsfonts}
\usepackage{algorithmic}
\usepackage{graphicx}
\usepackage{textcomp}
\usepackage{xcolor}
\usepackage{url}

\usepackage{booktabs}
\usepackage{array}

\usepackage{bm}
\makeatletter
\AtBeginDocument{\DeclareMathVersion{bold}
\SetSymbolFont{operators}{bold}{T1}{times}{b}{n}
\SetSymbolFont{NewLetters}{bold}{T1}{times}{b}{it}
\SetMathAlphabet{\mathrm}{bold}{T1}{times}{b}{n}
\SetMathAlphabet{\mathit}{bold}{T1}{times}{b}{it}
\SetMathAlphabet{\mathbf}{bold}{T1}{times}{b}{n}
\SetMathAlphabet{\mathtt}{bold}{OT1}{pcr}{b}{n}
\SetSymbolFont{symbols}{bold}{OMS}{cmsy}{b}{n}
\renewcommand\boldmath{\@nomath\boldmath\mathversion{bold}}}
\makeatother

\def\BibTeX{{\rm B\kern-.05em{\sc i\kern-.025em b}\kern-.08em
    T\kern-.1667em\lower.7ex\hbox{E}\kern-.125emX}}

\begin{document}
\history{Date of publication xxxx 00, 0000, date of current version xxxx 00, 0000.}
\doi{10.1109/ACCESS.2017.DOI}

\title{Trajectory-based actuator identification via differentiable simulation}

\author{\uppercase{Vyacheslav Kovalev}\authorrefmark{1},
\uppercase{Ekaterina Chaikovskaia}\authorrefmark{1},
\uppercase{Egor Davydenko}\authorrefmark{1}, and
\uppercase{Roman Gorbachev}\authorrefmark{1}}

\address[1]{Moscow Institute of Physics and Technology (MIPT), Dolgoprudny, Moscow Region, Russia}

\tfootnote{The study was supported by the Ministry of Economic Development of the Russian Federation (agreement with MIPT No.\ 139-15-2025-013, dated June 20, 2025, IGK 000000C313925P4B0002).}

\markboth
{Kovalev \headeretal: Trajectory-based Actuator Identification via Differentiable Simulation}
{Kovalev \headeretal: Trajectory-based Actuator Identification via Differentiable Simulation}

\corresp{Corresponding author: Vyacheslav Kovalev (e-mail: kovalev.vv@phystech.edu).}

\begin{abstract}
Accurate actuation models are critical for bridging the gap between simulation and real robot behavior, yet obtaining high-fidelity actuator dynamics typically requires dedicated test stands and torque sensing. We present a trajectory-based actuator identification method that uses differentiable simulation to fit system-level actuator models from encoder motion alone. Identification is posed as a trajectory-matching problem: given commanded joint positions and measured joint angles and velocities, we optimize actuator and simulator parameters by backpropagating through the simulator, without torque sensors, current/voltage measurements, or access to embedded motor-control internals. The framework supports multiple model classes, ranging from compact structured parameterizations to neural actuator mappings, within a unified optimization pipeline. On held-out real-robot trajectories for a high-gear-ratio actuator with an embedded PD controller, the proposed torque-sensor-free identification achieves much tighter trajectory alignment than a supervised stand-trained baseline dominated by steady-state data, reducing mean absolute position error from 14.20~mrad to as low as 7.54~mrad (1.88$\times$). Finally, we demonstrate downstream impact for the same actuator class in a real-robot locomotion study: training policies with the refined actuator model increases travel distance by 46\% and reduces rotational deviation by 75\% relative to the baseline.
\end{abstract}


\titlepgskip=-21pt

\maketitle

\section{Introduction}
\label{sec:introduction}

\subsection{Motivation: Actuator Modeling in Simulation for Accurate Robot Control}

Accurate modeling of a specific robot’s digital twin—i.e., the simulation model that stands in for the physical system—is foundational for reliable control, planning, and analysis \cite{di2018dynamic,kim2019highly,gaertner2021collision}.
By ``modeling'' we mean both (i) \emph{static} elements such as link masses and inertias, joint limits, kinematic geometry, and contact/friction parameters, and (ii) \emph{dynamic} processes that mediate commands into motion, most notably the actuation stack and embedded low‑level control (delays, bandwidth limits, saturations, nonlinearities).
Errors in either class degrade closed‑loop behavior and prediction fidelity across application areas including manipulation, locomotion, and mobile navigation.

In modern practice, many controllers are developed and tuned in simulation and then executed on hardware. This is true for model‑based designs as well as learning‑based controllers (including reinforcement learning (RL) formulations) \cite{hwangbo2019learning}.
The effectiveness of this workflow hinges on how closely the simulated robot reproduces the real one: mismatches in dynamics lead to performance drops at deployment.
For instance, in robotic grasping, controllers that succeed in simulation often suffer a marked decrease in success on real hardware due to unmodeled effects and distribution shift \cite{james2019sim}.

A primary source of mismatch is the \emph{actuation interface}—the final pathway from a controller’s command to joint torque and motion \cite{aljalbout2025reality}.
Even with accurate perception and rigid‑body dynamics, delays, bandwidth limits, nonlinearities, and command shaping by hidden low‑level control loops and safety filters can dominate observed behavior (the \emph{actuation gap}).
Unlike rigid‑body parameters, actuator behavior couples electrical, control, and mechanical effects that are difficult to capture with a single idealized model; consequently, actuator inaccuracies frequently limit control performance and sim‑to‑real transfer.

Most robotics simulators and control frameworks therefore simplify actuation.
A broad literature proposes complementary remedies ranging from structured actuator modeling and experimental characterization to robustness techniques (e.g., domain randomization) and trajectory‑matching identification, including recent differentiable‑simulation‑based methods.
We review these approaches and their trade‑offs in Section~\ref{sec:related_work}, and next outline our trajectory‑based identification method.

\subsection{Overview of the Proposed Approach}

In this work, we propose a trajectory-based actuator identification framework that uses \emph{differentiable} simulation to identify the overall behavior of robotic actuators directly from observable motion data in a model-agnostic manner.
We formulate identification as a trajectory-matching problem and solve it via gradient-based optimization enabled by simulator differentiability, without relying on torque sensors, internal electrical measurements, or explicit low-level motor models.
In our experiments, observable motion data consist of joint angles and velocities from encoders; the control input is the desired joint position.

The identified actuator models are validated through quantitative comparison with a supervised torque model trained on dedicated motor test-stand measurements, highlighting the limitations of steady-state characterization.
Furthermore, the proposed method captures actuator behavior beyond steady-state regimes, enabling tighter alignment between simulated and real systems under dynamic and interaction-rich conditions. While the framework is not tied in principle to a single actuator implementation, the experimental validation in this paper is limited to high-gear-ratio actuators with embedded PD control.

To connect single-actuator identification to system-level control, we then \emph{evaluate} the identified actuator model inside a locomotion training pipeline. Concretely, we train a simulation policy using the refined actuator parameters and compare it against a baseline policy trained with an unmodified (default) actuator model, mirroring common practice when accurate actuator parameters are unavailable.
Finally, the refined actuator model provides supporting downstream validation:
in a real-robot locomotion study (Section~\ref{sec:scope_applications}), a policy trained with the refined motor model traveled farther and showed less rotation, with travel distance increasing by 46\% and rotational deviation decreasing by 75\% relative to the baseline (mean \(\pm\) std over 10 runs; Fig.~\ref{fig:rl_comparison}, Table~\ref{tab:results}), providing supporting downstream validation of the identified model.

\subsection{Contributions}
This paper makes the following contributions:
\begin{itemize}
    \item \textbf{Trajectory-based, torque-sensor-free identification:} a differentiable-simulation pipeline that identifies system-level actuator behavior from encoder motion alone (no torque or electrical sensing), avoiding test-stand instrumentation. The framework is model-agnostic in formulation, while the present experimental validation is on high-gear-ratio actuators with embedded PD control.

    \item \textbf{Unified differentiable optimization framework:} a trajectory-matching pipeline spanning (i) structured/parametric actuator mappings, (ii) neural actuator models, and (iii) physics-engine internal parameters (e.g., armature, joint friction, delay).

    \item \textbf{Torque-oracle identification mode:} directly estimates per-timestep torque (a reference for the tightest trajectory fit the framework can produce: with no parametric constraints on the torque sequence, it represents the limit of expressiveness within the same simulator and optimizer), yielding pseudo-labels for subsequent supervised training without a torque sensor.

     \item \textbf{Reproducibility (code):} we provide a publicly accessible repository containing the main identification and evaluation code, one example dataset, and example scripts/instructions for reproducing representative results. Repository link: \url{https://wavegit.mipt.ru/Slavoch/mjx_sysid}.

    \item \textbf{Quantitative validation on held-out trajectories:} under identical commands, compared to a supervised test-stand-trained baseline (mean absolute position error, MAE \(=\)14.20~mrad), our torque-sensor-free trajectory identification achieves MAE as low as 7.54~mrad (\(\approx\)1.88\(\times\) lower), demonstrating tighter trajectory alignment without test-stand hardware.

    \item \textbf{Downstream impact on real-robot locomotion:} in a real-robot study (Section~\ref{sec:scope_applications}), a policy trained with the refined motor model shows a consistent practical advantage over the default baseline, with mean travel distance increasing by 46\% and mean rotational deviation decreasing by 75\% over 10 runs (Fig.~\ref{fig:rl_comparison}, Table~\ref{tab:results}).
\end{itemize} 

\subsection{Paper organization}
Section~\ref{sec:related_work} reviews actuator modeling and compensation approaches.
Section~\ref{sec:method} presents our trajectory-based actuator identification via differentiable simulation and gradient-based optimization of actuator parameters.
Section~\ref{sec:validation} presents experimental results and baseline comparisons, followed by scope and downstream applications (Section~\ref{sec:scope_applications}) and conclusions (Section~\ref{sec:conclusion}).

\section{Related Work and Comparative Background}
\label{sec:related_work}

Actuator modeling and compensation in robotics spans three principal families (Table~\ref{tab:actuator_modeling_comparison}): (i) simplified actuator abstractions used for control and simulation (Section~\ref{subsec:actuator_abstractions}); (ii) hardware characterization on dedicated test stands (Section~\ref{subsec:hardware_characterization}); and (iii) robustness‑ and learning‑driven strategies, including domain randomization, offline trajectory‑matching identification, and neural residual compensators (Sections~\ref{subsec:domain_randomization}, \ref{subsec:offline_traj_id}, \ref{subsec:neural_residuals}).

Table~\ref{tab:actuator_modeling_comparison} uses qualitative \emph{low/high} labels to convey two axes.
\emph{Model complexity} is a proxy for the number of trainable parameters and runtime evaluation cost: compact parametric models (e.g., proportional–derivative (PD)‑like mappings with few parameters) are \emph{low} complexity, whereas neural actuator models (neural-network-based, many parameters) are \emph{high} complexity.
\emph{Interpretability} follows the inverse trend: low‑parameter, structured models are typically more transparent and thus \emph{high} interpretability, while high‑capacity neural models are typically less transparent and thus \emph{low} interpretability.

Each family makes distinct trade‑offs among physical fidelity, instrumentation/data requirements, and ease of deployment. We focus on learning \emph{transient, system‑level} actuator behavior \emph{without} torque sensors or test-stand hardware via trajectory‑based identification with differentiable simulation, and refer back to Table~\ref{tab:actuator_modeling_comparison} for a side‑by‑side view.

\begin{table*}[t]
\caption{\textbf{Comparison of actuator modeling and compensation approaches.} Model complexity is a qualitative proxy for the number of trainable parameters and runtime evaluation cost.}
\label{tab:actuator_modeling_comparison}
\centering
\footnotesize
\setlength{\tabcolsep}{2pt}
\renewcommand{\arraystretch}{1.1}
\begin{tabular}{p{0.18\textwidth} p{0.16\textwidth} p{0.15\textwidth} p{0.15\textwidth} p{0.15\textwidth} p{0.15\textwidth}}
\hline
Approach & Instrumentation & Data Regime & Optimization & Interpretability & Model complexity \\
\hline
Ideal PD/torque abstraction & None & Simulation assumption & — & High & Low \\
\hline
Domain randomization & None & Sim randomized & — & — & — \\
\hline
Test-stand characterization & Torque sensor + test stand & Test stand (primarily quasi-static) & Supervised & Low-High & Low-High \\
\hline
Offline traj-ID & None & Real rollouts & Self-supervised; gradient-free & High & Low \\
\hline
Learned actuator modules / residual action & None & Real rollouts & Self-supervised; gradient-based & Low & High \\
\hline
Differentiable traj-ID (ours) & None & Real rollouts & Self-supervised; gradient-based & Low--High & Low--High \\
\hline
\end{tabular}
\end{table*}

\subsection{Actuator Modeling Abstractions in Robotics}
\label{subsec:actuator_abstractions}

A standard abstraction treats joints as near‑ideal torque sources shaped by PD/impedance laws \cite{xie2023learning,siekmann2021sim,rodriguez2021deepwalk,li2024reinforcement,ravichandar2025preferenced,zhang2025natural,rudin2022learning,makoviychuk2021isaac,mittal2025isaaclab}. This enables practical design and fast simulation but collapses inner‑loop control, electronics, and electrical dynamics into an idealized mapping.

Despite its convenience, the torque-source abstraction neglects several phenomena that strongly influence real actuator behavior.
Electrical dynamics introduce delays and bandwidth limitations, while inverter nonlinearities, dead-time effects, and current saturation distort the relationship between commanded and realized torque \cite{schmidt2022practical}.
Mechanical effects (e.g., friction/backlash) vary with load, configuration, and motion \cite{bittencourt2010extended,wolf2018extending}, further widening the abstraction gap.
Low-level field-oriented control, direct torque control, and model predictive control regulate currents to approximate linear torque over limited bands \cite{wang2018advanced,zhang2016overview}, yet system‑level behavior remains the combination of electronics, inner loops, and uncertainties—hard to capture with a single simplified model.

Consequently, even when advanced low-level motor control is used, the torque-source abstraction remains an approximation.
This motivates system identification approaches that aim to capture the effective actuator behavior as experienced by high-level controllers, rather than attempting to model each internal component explicitly.

\subsection{Hardware-Based Actuator Characterization}
\label{subsec:hardware_characterization}

A traditional and widely adopted approach to improving actuator models relies on experimental characterization using dedicated motor or actuator test stands.
By mechanically decoupling the actuator from the robot and subjecting it to controlled excitation and loading conditions, test stands provide a means to directly measure actuator behavior under well-defined operating conditions.
These setups often incorporate torque transducers (either inline or reaction torque types), dynamometers, and synchronized measurement of speed and electrical quantities to directly capture mechanical output across operating points \cite{MARTYR2007144,sziki2022measurement,lee2023performance}.
Motor test stands have been developed specifically to obtain key performance metrics such as torque, speed, efficiency, and dynamic response that are difficult to estimate accurately from electrical measurements or manufacturer specifications alone \cite{MARTYR2007144}.
Such empirical data from test stands has been widely used in both industrial and academic contexts to refine actuator models and validate simulation-based predictions \cite{sziki2022measurement,lee2023performance}.

Test stands typically allow precise control of rotational speed, load torque, or both, while directly measuring mechanical quantities such as output torque and angular velocity.
This setup enables identification of steady-state actuator characteristics, including torque--speed curves, efficiency maps, friction parameters, and effective torque constants.
Such measurements are commonly used to calibrate simulation models and validate analytical assumptions.

Torque measurement in actuator test stands is most commonly realized using either inline torque sensors or reaction torque measurement.
Inline torque sensors directly measure transmitted shaft torque through torsional strain, providing high bandwidth and sensitivity at the expense of added inertia, compliance, and mechanical integration complexity.
Reaction torque measurement infers torque from the force exerted on a mechanically constrained motor or brake housing and offers a simpler and more robust mechanical design.
Most rigs target steady-state conditions for reliable mapping and repeatability.

Under steady-state operation at constant angular velocity, inertial effects vanish and the rotational dynamics become a static torque balance.
In this case, measured torque corresponds directly to the average output torque of the actuator, enabling data-based identification of mappings between control commands, speed, and torque.
These mappings in practice capture combined effects of electrical dynamics, control nonlinearities, and mechanical losses.

Despite their practical value, hardware-based characterization methods have important limitations.
Steady-state measurements do not capture transient dynamics, inertial effects, or high-frequency torque variations that occur during rapid accelerations, impacts, or contact events.
Moreover, test-stand experiments require specialized hardware, careful mechanical setup, and dedicated calibration procedures, which can be costly and difficult to scale across large robotic platforms.

Nevertheless, test-stand-based actuator characterization remains a valuable tool and is widely used in both academic and industrial settings.
It provides a clear improvement over idealized torque-source assumptions and serves as a reference point for evaluating alternative identification methods.
In this work, test-stand measurements are used as a comparative baseline to highlight the limitations of steady-state characterization and to motivate trajectory-based identification approaches that capture actuator behavior under dynamic operating conditions.

\subsection{Domain Randomization}
\label{subsec:domain_randomization}
Domain randomization perturbs key simulation parameters (e.g., mass, friction, actuator gains/delays, torque limits) during training to encourage robustness \cite{li2024reinforcement,akkaya2019solving,tiboni2023domain}.

In practice, domain randomization encourages robustness primarily with respect to the \emph{chosen} training distribution; there is no guarantee that the real system lies within (or is well covered by) the randomized parameter ranges.
Consequently, performance can be highly sensitive to which parameters are randomized and how their ranges are selected, which is typically problem-dependent and requires manual tuning.
If the randomization is too broad, the induced uncertainty can bias learning toward risk-averse (over-conservative) behaviors and can slow training or degrade asymptotic performance, especially in high-dimensional settings.
Empirically, overly ``heavy'' randomization may underperform more targeted or ``mild'' randomization \cite{james2019sim}.

\subsection{Offline Trajectory-Matching Identification}
\label{subsec:offline_traj_id}
More recently, learning-based approaches have explored identifying actuator dynamics directly from trajectory data by fitting model parameters such that simulated trajectories match real-world observations under identical control inputs.
In practice, this is often performed using controlled excitation signals and gradient-free optimization (e.g., evolution strategies) when the simulator is non-differentiable \cite{haarnoja2024learning,duclusaud2025extended}.
The main advantages are practicality and instrumentation simplicity: these methods can rely on observable kinematics (joint positions/velocities) without torque sensing or access to internal electrical states; however, they can require carefully designed excitation and may be sample-inefficient or less reliable outside the identification setting.

\paragraph{Limitation.} Gradient-free search can be sample-inefficient and sensitive to excitation design; when gradients are available, identification often converges faster and more reliably.

\subsection{Differentiable Trajectory-Matching Identification}
\label{subsec:diff_traj_id}
Differentiable simulators enable gradient-based system identification by backpropagating trajectory-matching losses through dynamics \cite{brax2021github,heiden2021neuralsim}. This makes it possible to fit actuator- and simulator-level parameters (e.g., gains, friction, delay, and contact-related terms) directly from rollouts, often improving optimization efficiency compared to gradient-free search.

Among widely used robotics simulators, MuJoCo is frequently reported to provide strong accuracy and numerical stability across diverse tasks \cite{erez2015simulation,liao2023performance} and remains a common platform for RL and control research \cite{xie2018feedback,xie2019iterative,li2024reinforcement}. MJX (the JAX-based, differentiable MuJoCo) further enables end-to-end gradient-based fitting and is a widely adopted differentiable simulator in modern RL workflows \cite{kaup2024review}.

Building on these foundations, our contribution is a scalable, torque-sensor-free trajectory-matching identification method that leverages differentiable simulation to recover \emph{system-level} actuator behavior from observable motion alone, and is validated in real-robot experiments.

In summary, prior approaches trade off instrumentation, optimization efficiency, interpretability, and fidelity. Our work targets the gap of \emph{transient, system-level} actuator identification \emph{without} torque sensors or benches, using differentiable trajectory matching to obtain compact, physically meaningful actuator models that can be validated quantitatively and translated into downstream control gains.

\subsection{Neural Residual Actuation Compensation}
\label{subsec:neural_residuals}
Several works learn neural modules to compensate dynamics mismatch—including actuation effects—so simulated rollouts align more closely with real trajectories \cite{fey2025bridging,he2025asap}.
In such approaches, the model (or simulator parameters) is trained to reduce state-prediction error so that rollouts more closely match measured real-system trajectories.
For example, Fey \emph{et al.}~\cite{fey2025bridging} introduce an \emph{Unsupervised Actuator Net} (UAN), a learned actuator module trained from real-world data without requiring torque sensing, and use it within a two-stage RL pipeline to improve sim-to-real transfer for athletic loco-manipulation.
He \emph{et al.}~\cite{he2025asap} propose ASAP, which deploys a pre-trained motion-tracking policy to collect real-world rollouts and then learns a residual (\emph{delta}) action model that compensates for dynamics mismatch; this model is integrated into the simulator for subsequent policy fine-tuning.
Compared to such neural residual compensators, our trajectory-based differentiable identification targets a compact and physically interpretable actuator parameterization that can be directly validated and stress-tested, at the cost of reduced expressiveness for unmodeled, history-dependent effects.

The main advantage of these learning-based models is expressiveness: in principle, complex and history-dependent effects can be captured when sufficient excitation and data are available.
However, purely data-driven models may be less physically interpretable and can be difficult to validate outside the training distribution.

\paragraph{Limitation.} Residual models can overfit to specific data regimes and reduce interpretability; validation outside the training distribution is challenging. In contrast, our parameterization emphasizes physical interpretability and direct validation at the actuator/system level.

\section{Proposed Method: Trajectory-based Actuator Identification via Differentiable Simulation}
\label{sec:method}

\begin{figure}[t]
    \centering
    \includegraphics[width=3in]{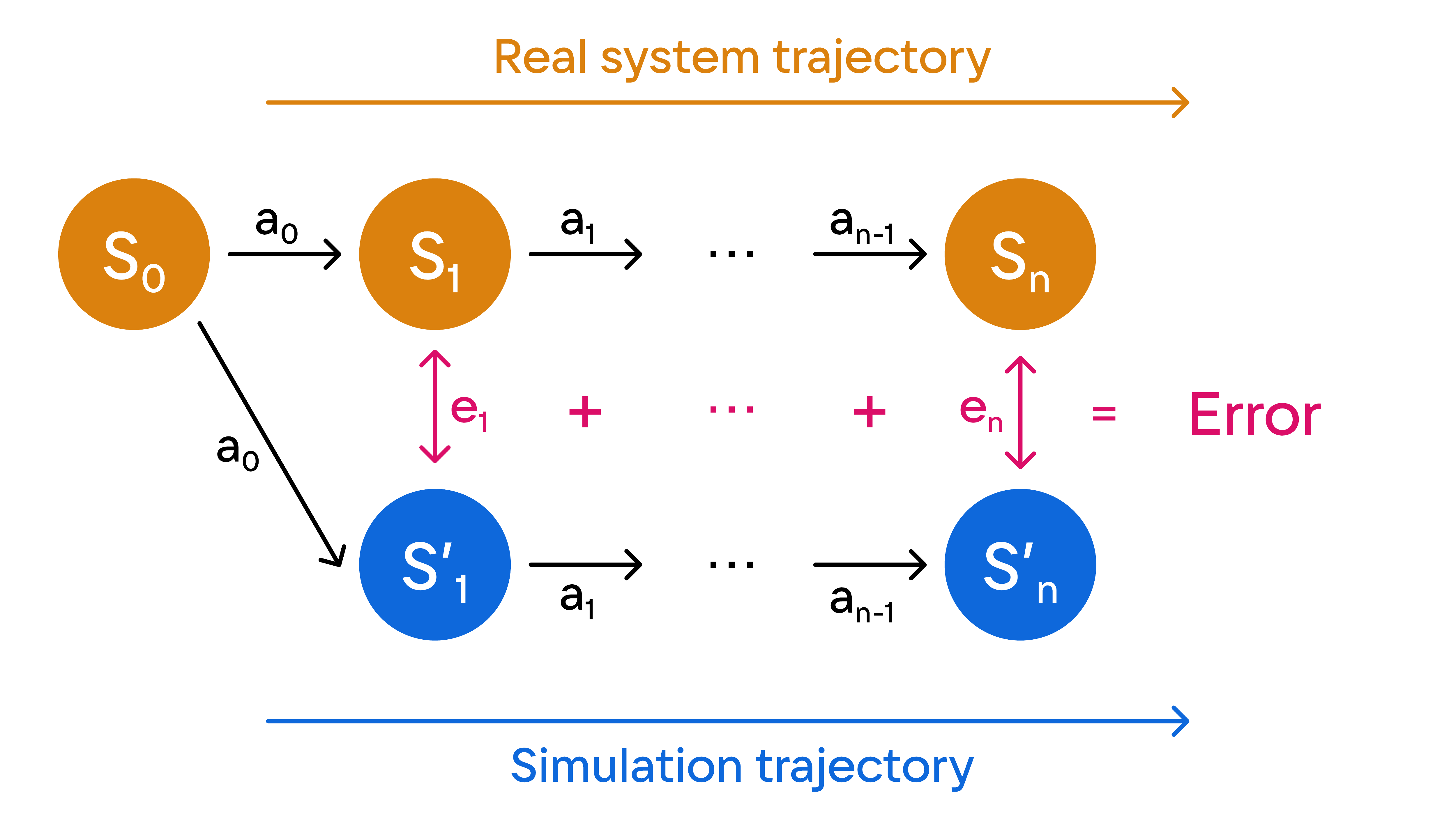}
    \caption{
        \textbf{Illustration of the trajectory-matching optimization:}
        The real system follows a trajectory $\{s_i\}$ under control inputs $\{a_i\}$.
        Simulated states $\{s'_i\}$ are generated using a parameterized simulator.
        At each timestep, the state prediction error \(e_i := s'_i - s_i\) is computed, and the total trajectory error is minimized during optimization.
    }
    \label{fig:problem_statement}
\end{figure}

We pose actuator identification as a trajectory-matching problem and solve it with gradient-based optimization enabled by a differentiable simulator (MJX). Consider a system with state $s_i$ and control input $a_i$ at timestep $i$. We assume the existence of an unknown parameter vector $z^*$ (e.g., actuator and/or simulator parameters) that, if known, would allow the simulator to reproduce the real system dynamics:
\begin{equation}
    s_{i+1} = s'_{i+1} = \Phi_{z^*}(s_i, a_i, \delta),
    \label{eq:next_state}
\end{equation}
where $s'_{i+1}$ is the predicted next state, $\delta$ is the simulation timestep, and $\Phi_z(s, a, \delta)$ denotes a discrete-time integration of the system dynamics parameterized by $z$. The timestep $\delta$ should be chosen small enough to achieve the desired numerical accuracy while avoiding excessive computational cost~\cite{aeran2016time}. Equation~\eqref{eq:next_state} expresses the ideal condition in which the simulator perfectly matches the real system.

In practice, we identify $z$ by minimizing a trajectory-matching loss over measured rollouts, as illustrated in Fig.~\ref{fig:problem_statement}.

\paragraph{System and parameterization.}
In our experiments the measured state is \(s=[q,\dot q]\), where \(q\) is joint angle and \(\dot q\) is joint velocity from encoders.
Both components are retained in the observed state for generality and for simulator rollout dynamics. However, their contributions to the training loss are controlled separately by the weight matrix \(W\) in Eq.~\eqref{eq:final_opt_problem}.
The control input is the commanded joint position \(a=q^{\mathrm{des}}\) produced by the higher‑level controller.
The parameter vector \(z\) collects actuator/simulator parameters and depends on the chosen model.
This might be a subset of MuJoCo joint parameters (e.g., armature, damping, frictionloss) or neural network weights.
The exact choice of \(z\) for each model is detailed in Section~\ref{subsec:model_learning}.
All parameters have physical bounds.

\subsection{Objective}
We minimize a weighted output mismatch without any parameter regularization to avoid biasing physically meaningful parameters (e.g., PD gains).
Numerical integration errors can accumulate over long horizons, potentially degrading the optimization. To mitigate this effect, the trajectory is divided into short segments of length $N$, and each optimization step minimizes the following minibatch objective over $M$ sampled segments:
\begin{equation}
    \begin{aligned}
        \mathcal{L}_{\mathrm{batch}}(z) = & \frac{1}{MN}\sum_{j=0}^{M-1}\sum_{i=1}^{N}\left\|W\big(s'_{i,j}-s_{i,j}\big)\right\|_2^2 \\
        \text{s.t.}\quad & s'_{0,j} = s_{0,j}, \\
        & s'_{i+1,j} = \Phi_z(s'_{i,j}, a_{i,j}, \delta),
    \end{aligned}
    \label{eq:final_opt_problem}
\end{equation}
where \(W=\mathrm{diag}(w_q,w_{\dot q})\) is a diagonal weight matrix penalizing angle and velocity residuals.
Thus, state components may be present in the rollout state and model inputs even when their residuals are not penalized in the loss.
Here, \(i\in\{1,\dots,N\}\) indexes timesteps within a segment and \(j\in\{0,\dots,M-1\}\) indexes sampled segments.
Each sampled segment \(j\) is rolled out from the measured initial state \(s'_{0,j}=s_{0,j}\).
Windows are extracted with overlap.
The segmentation policy and concrete values of \(N\), \(M\), and \(\delta\) are specified in Appendix~\ref{sec:our_apprach_appendix}.

\subsection{Limitations and Assumptions}

This formulation assumes that the primary discrepancy between simulation and reality can be captured by the parameter vector $z$. If unmodeled effects—such as external disturbances, time-varying properties, or sensor noise—are significant and not included in $z$, the predicted states $s'_{i+1}$ may deviate from the true dynamics.

Moreover, achieving exact equality in \eqref{eq:next_state} is generally impractical due to inherent modeling complexity. Nonetheless, the approach remains effective as long as the simulator with optimized parameters $\Phi_{z^*}(s'_i, a_i, \delta)$ provides an accurate approximation of the true system dynamics within an acceptable error margin.

\section{Validation and Comparative Analysis}
\label{sec:validation}
We evaluate on a motor from the ROKI-2 robot~\cite{starkit_roki2}; hardware details are provided in Appendix~\ref{app:bench}.
Our main experimental validation focuses on a high-gear-ratio actuator (210:1) with a hand‑tuned embedded controller, a setting whose dynamics are challenging to reproduce in simulation.
Our objective is to reproduce the real robot’s joint motion as accurately as possible in simulation.
We train three actuator models of increasing complexity using the proposed trajectory‑matching identification procedure.
As baselines, we include a supervised model trained on test-stand torque data and a gradient-free black-box optimization model.

Model parameterizations and training settings are summarized in Section~\ref{subsec:model_learning}.
We then present the trajectory‑matching experiment and comparative results in Section~\ref{subsec:exp_datasets}.
Optimization stability and parameter identifiability are assessed in Section~\ref{subsec:stability_sensitivity}.
Sensitivity to the objective weighting matrix is analyzed in Section~\ref{subsec:w_sensitivity}, and the effect of prediction horizon is studied in Section~\ref{subsec:segment_length_ablation}.
Conclusions and broader implications are discussed in Section~\ref{sec:conclusion}.

\begin{table}[bh]
\centering
\caption{\textbf{Comparison} of actuator models and training protocol.}
\label{tab:models_protocol}
\small
\setlength{\tabcolsep}{5pt}
\renewcommand{\arraystretch}{1.15}
\begin{tabular}{l p{0.30\linewidth} p{0.3\linewidth}}
\toprule
\textbf{Model} & \textbf{Supervision / data} & \textbf{Model type} \\
\midrule
\multicolumn{3}{l}{\textbf{Stand-based supervised baseline}} \\
\textsc{Bench-Sup} &
Supervised. &
MLP \\
\midrule
\multicolumn{3}{l}{\textbf{Trajectory-based gradient-free baseline}} \\
\textsc{Param-ES} &
Self-supervised; gradient-free.&
PD gains, armature (evolution strategy) \\
\midrule
\multicolumn{3}{l}{\textbf{Trajectory-based residual-action (ASAP) baseline}} \\
\textsc{Residual-RL} &
Self-supervised.&
MLP \\
\midrule
\multicolumn{3}{l}{\textbf{Trajectory-based, torque-sensor-free (proposed, \(\eqref{eq:final_opt_problem}\))}} \\
\textsc{TrajID-Param} &
Self-supervised. &
PD gains, armature \\
\textsc{TrajID-NN} &
Self-supervised. &
MLP \\
\textsc{Torque-Oracle} &
Self-supervised. &
Torque sequence \(\{\tau_i\}_{i=1}^{N}\)\\
\bottomrule
\end{tabular}
\end{table}

\subsection{Model Learning}
\label{subsec:model_learning}
Table~\ref{tab:models_protocol} summarizes the compared models and training protocol.
All torque-model baselines share the input–output signature \((q^{\mathrm{des}},\, q,\, \dot q)\!\mapsto\!\tau\), where \(\tau\in\mathbb{R}\) denotes joint torque.
The ASAP-inspired baseline \textsc{Residual-RL} instead learns a residual command correction \(\delta a\) that is added to the desired position command before simulation.

All non‑baseline models are trained via our trajectory-matching identification pipeline without torque sensing, using robot rollouts under known loads.
This yields datasets of \((q^{\mathrm{des}}, q, \dot q)\) sufficient to fit actuator parameters that reproduce measured trajectories.
Dataset design and optimization hyperparameters are detailed in Appendix~\ref{sec:our_apprach_appendix}.
Unless otherwise noted, optimization settings are held constant across models for a fair comparison.
Training minimizes the trajectory‑matching objective in Eq.~\eqref{eq:final_opt_problem}.

\begin{itemize}
    \item \textbf{Baseline (\textsc{Bench-Sup}):}
    a supervised multilayer perceptron (MLP) mapping \((q^{\mathrm{des}}, q, \dot q)\!\to\!\tau\), trained on ground‑truth torque from a motor test‑stand dataset; architectural and data‑collection details are provided in Appendix~\ref{app:bench}.
    \item \textbf{Baseline (\textsc{Param-ES}):}
    the same parametric model family as \textsc{TrajID-Param} (PD gains, armature), but optimized with sMAES~\cite{hansen2016cma}, a population-based derivative-free evolution strategy, as a gradient-free baseline.
    \item \textbf{Baseline (\textsc{Residual-RL}):}
    an ASAP-inspired residual-action baseline that learns a correction \(\delta a_i\) to the commanded position rather than actuator parameters. Unlike the original ASAP pipeline, this proxy is implemented in the same simulator environment as the other compared models for fairness. Because it operates in command space, it does not directly update built-in simulator parameters such as armature or friction loss. For comparability, we match the hidden-layer size of \textsc{TrajID-NN}, train on the same rollout data budget, and evaluate it with the same held-out rollout metric; training details are given in Appendix~\ref{sec:our_apprach_appendix}.
    \item \textbf{Identified parametric TrajID (\textsc{TrajID-Param}):}
    a deliberately simple, interpretable actuator model that optimizes PD gains (\(k_p\): position stiffness, \(k_v\): velocity damping) together with a MuJoCo joint parameter (armature)—three parameters in total—to match measured trajectories.
    \item \textbf{Identified neural TrajID (\textsc{TrajID-NN}):}
    a feed‑forward MLP, used to capture richer static nonlinearities.
    For fairness, it uses the same instantaneous inputs (no recurrence, no state history). This design keeps the comparison aligned with the other actuator models, but it also limits the network's ability to represent time-dependent effects. Model size and training details are given in Appendix~\ref{sec:our_apprach_appendix}.
    \item \textbf{Torque oracle (\textsc{Torque-Oracle}):}
    an upper‑bound reference that directly optimizes a free per‑timestep torque sequence \(\{\tau_i\}_{i=1}^{N}\) (i.e., \(N\) free parameters per evaluated trajectory segment), indicating the best achievable trajectory fit; it also yields pseudo‑labels \((q^{\mathrm{des}}, q, \dot q, \tau)\) that can seed subsequent supervised training.
\end{itemize}

\subsection{Experiment}
\label{subsec:exp_datasets}

\begin{figure}[!t]
\centering
\includegraphics[width=\linewidth]{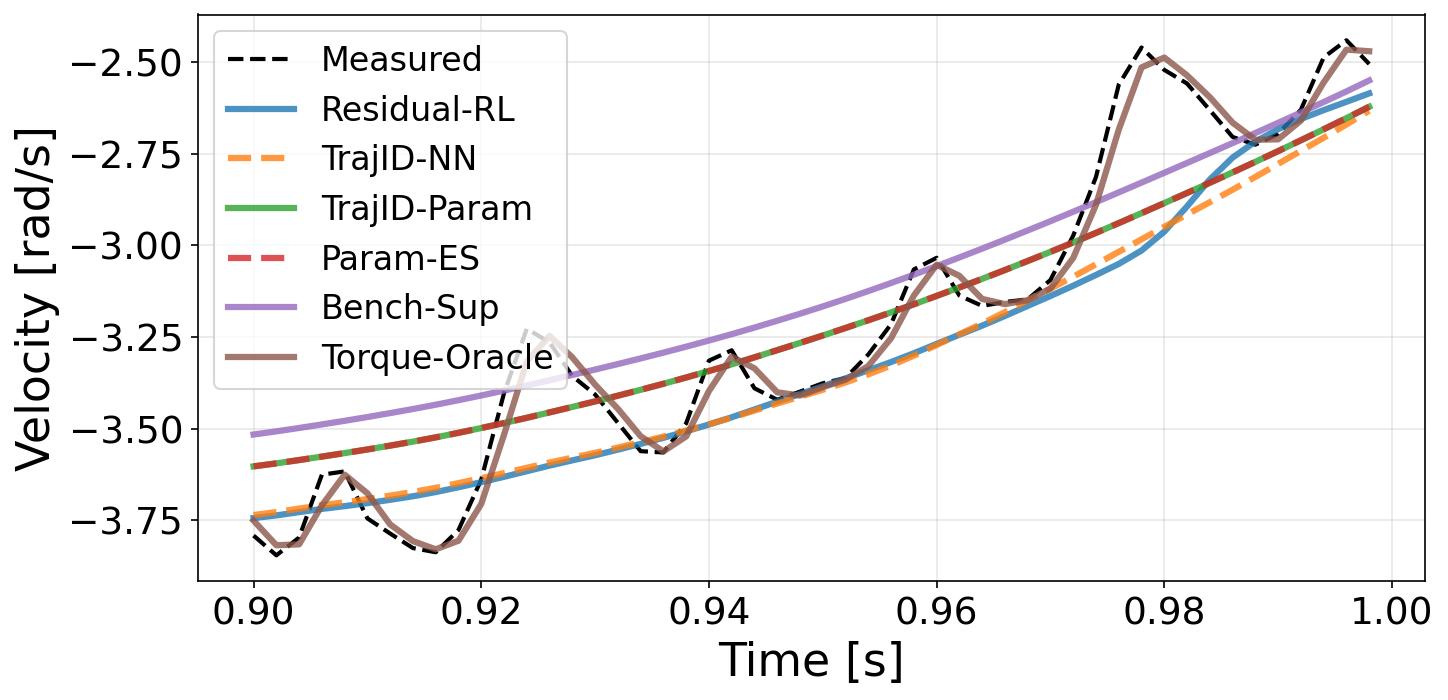}
\caption{\textbf{Measured vs simulated motor response under identical control inputs.} Motor velocity measured on hardware (black dashed) versus simulated trajectories for each model (colored lines).}
\label{fig:traj_example}
\end{figure}

\begin{figure}[!t]
\centering
\includegraphics[width=\linewidth]{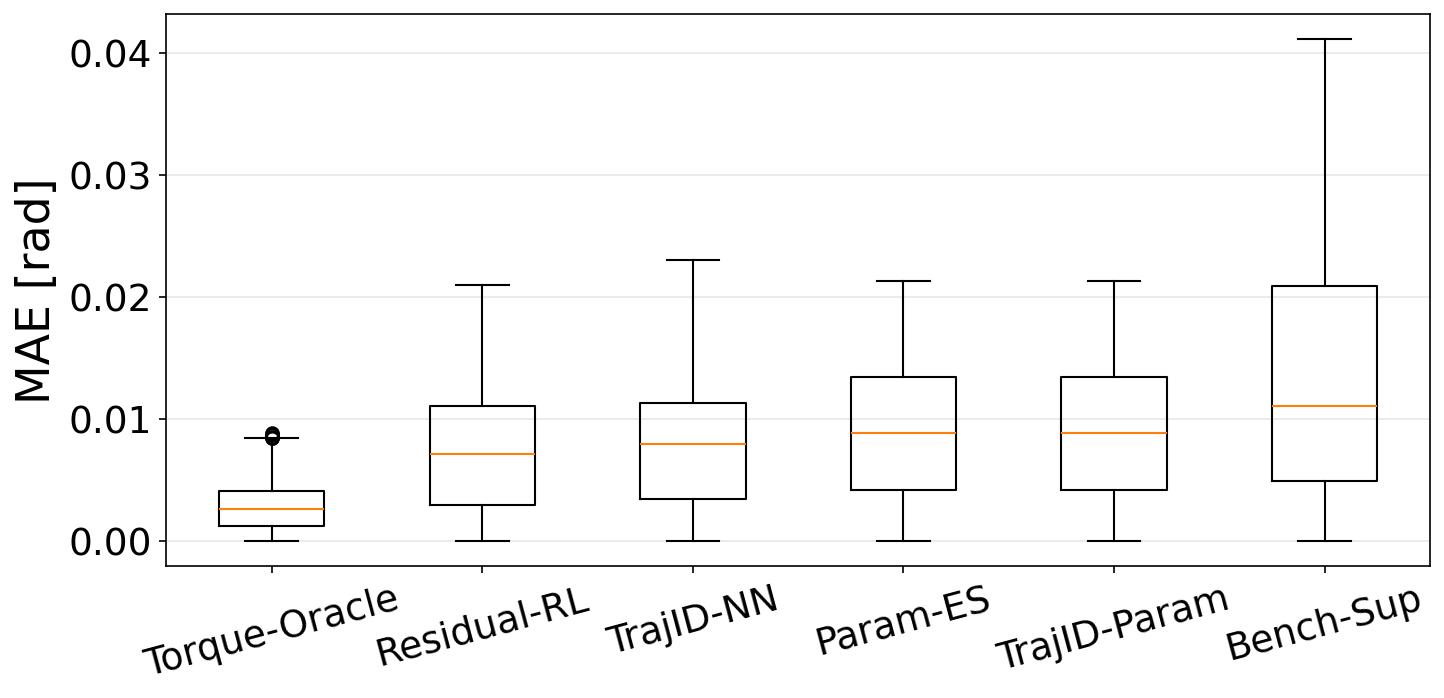}
\caption{\textbf{Position MAE comparison across actuator models.} Box-and-whisker plot of the mean absolute position error (Eq.~\ref{eq:mae}) on the held-out trajectory. Boxes indicate the interquartile range; the central line indicates the median; whiskers show the spread of the observed values.}
\label{fig:error_bars}
\end{figure}

\paragraph{Description.}
We evaluate how accurately each actuator model reproduces measured joint trajectories under identical command sequences.
Models are compared by their ability to minimize the discrepancy between simulated and measured motions in position (and velocity).
We collect time‑aligned joint position \(q\) and velocity \(\dot q\) under known loads using the same experimental protocol (sampling rate, command design, trajectories) described in Appendix~\ref{sec:our_apprach_appendix}.

\paragraph{Setup.}
Each rollout is initialized from the measured initial state \(s_0=[q_0,\dot q_0]\).
For the torque-model baselines, at every timestep \(i\), the model receives \((q^{\mathrm{des}}_i,\, q^{\mathrm{sim}}_i,\, \dot q^{\mathrm{sim}}_i)\) and outputs a torque \(\tau_i\); the simulator then advances the state.
For \textsc{Residual-RL}, the model outputs a residual command correction \(\delta a_i\), and the simulator is driven by the corrected command \(\tilde a_i=a_i+\delta a_i\).
No future or measured states beyond the initial condition are used.
Control and simulation run at 500\,Hz for 10\,s per trajectory.
We report trajectory‑tracking accuracy using the MAE:

\begin{equation}
\mathrm{MAE} = \frac{1}{H}\sum_{i=1}^{H}\lvert q^{\mathrm{sim}}_i - q^{\mathrm{meas}}_i\rvert.
\label{eq:mae}
\end{equation}
Here \(H\) is the total number of evaluation timesteps, \(q^{\mathrm{sim}}_i\) the simulated joint angle, and \(q^{\mathrm{meas}}_i\) the measured joint angle at step \(i\).

\paragraph{Results.}
Fig.~\ref{fig:traj_example} shows representative overlays of measured and simulated trajectories.
All models track joint position closely; differences are most visible in velocity during fast transients.

Quantitative results are shown in Fig.~\ref{fig:error_bars}.
The test-stand baseline \textsc{Bench-Sup} yields the highest error (\(14.20 \pm 11.50\)~mrad), reflecting its steady-state-only training data.
All trajectory-identified models outperform it substantially.

\textsc{Residual-RL} achieves \(7.17 \pm 4.56\)~mrad, i.e., nearly the same performance as \textsc{TrajID-NN}, which is expected because both use neural models with comparable capacity.
\textsc{TrajID-NN} achieves \(7.54 \pm 4.57\)~mrad, demonstrating that gradient-based differentiable optimization can exploit a higher-capacity model.
This comparison suggests that action-space residual compensation can match a neural torque model on this task when model capacity is similar. In this sense, \textsc{Residual-RL} covers the ASAP-style action-residual family. However, residual-action methods such as \textsc{Residual-RL} do not directly identify built-in simulator parameters such as armature, whereas our trajectory-identification framework can optimize those quantities and also supports compact analytical parameterizations such as \textsc{TrajID-Param}.
In addition, our approach is \(23.4\)\(\times\) faster to train in wall-clock time, requiring \(1.98\) minutes for \textsc{TrajID-NN} versus \(46.34\) minutes for \textsc{Residual-RL}.

\textsc{TrajID-Param} and \textsc{Param-ES} reach nearly identical errors (\(8.73 \pm 5.31\)~mrad each), showing that the gradient-free optimizer converges to the same solution on this low-dimensional (\(3\)-parameter) problem.
Both methods also converge in nearly the same wall-clock time, approximately \(1.43\) minutes each.
Crucially, the advantage of our gradient-based method becomes apparent with higher-capacity parameterizations: \textsc{TrajID-NN} requires optimizing over \({\sim}1200\) parameters, a regime where gradient-free search is impractical, yet differentiable simulation handles it directly.
\textsc{Torque-Oracle} serves as an upper bound (\(2.86 \pm 2.12\)~mrad), reflecting the best achievable fit with a free per-timestep torque sequence.

\subsection{Stability and Sensitivity Analysis}
\label{subsec:stability_sensitivity}
\begin{figure}[!t]
\centering
\includegraphics[width=\linewidth]{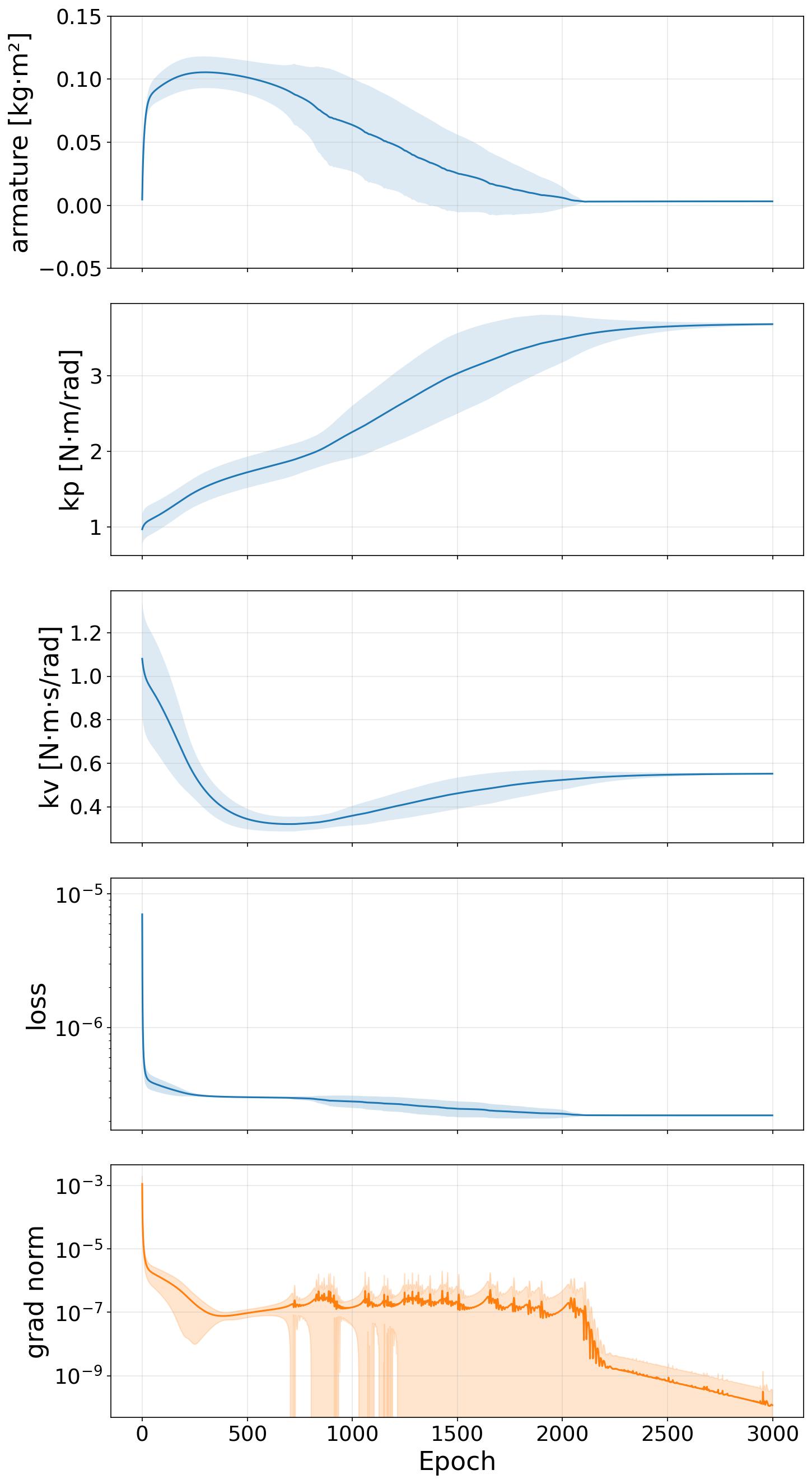}
\caption{Parameter, loss, and gradient norm convergence of \textsc{TrajID-Param} over 25 independent runs. Each run lasts 3000 epochs. Shaded regions indicate mean \(\pm\) std.}
\label{fig:stability_convergence}
\end{figure}

To assess optimization stability and parameter identifiability, we run \textsc{TrajID-Param} for 25 independent trials with different random seeds, varying both batch sampling and parameter initialization. Each run lasts 3000 epochs.
Fig.~\ref{fig:stability_convergence} shows the convergence of parameters, loss, and gradient norm across runs (mean \(\pm\) std).

The identified parameters are highly consistent across runs: \(k_p\) (position stiffness) \(= 3.684 \pm 0.013\)~N\,m/rad, \(k_v\) (velocity damping) \(= 0.552 \pm 0.002\)~N\,m\,s/rad, and armature \(= 0.00321 \pm 2.5\times10^{-5}\)~kg\,m\(^2\).
The best loss is \(2.211\times10^{-7} \pm 3.2\times10^{-12}\) across all runs. In every trial, training converged to this value without early stopping.

In summary, the near-zero dispersion in both parameters and loss suggests that the objective is well-conditioned and produces stable solutions in our experiments.

\subsection{Objective-Weight Sensitivity (\(W\) Sweep)}
\label{subsec:w_sensitivity}
\begin{figure}[!t]
\centering
\IfFileExists{experiment2_alpha_mae.jpg}{\includegraphics[width=\linewidth]{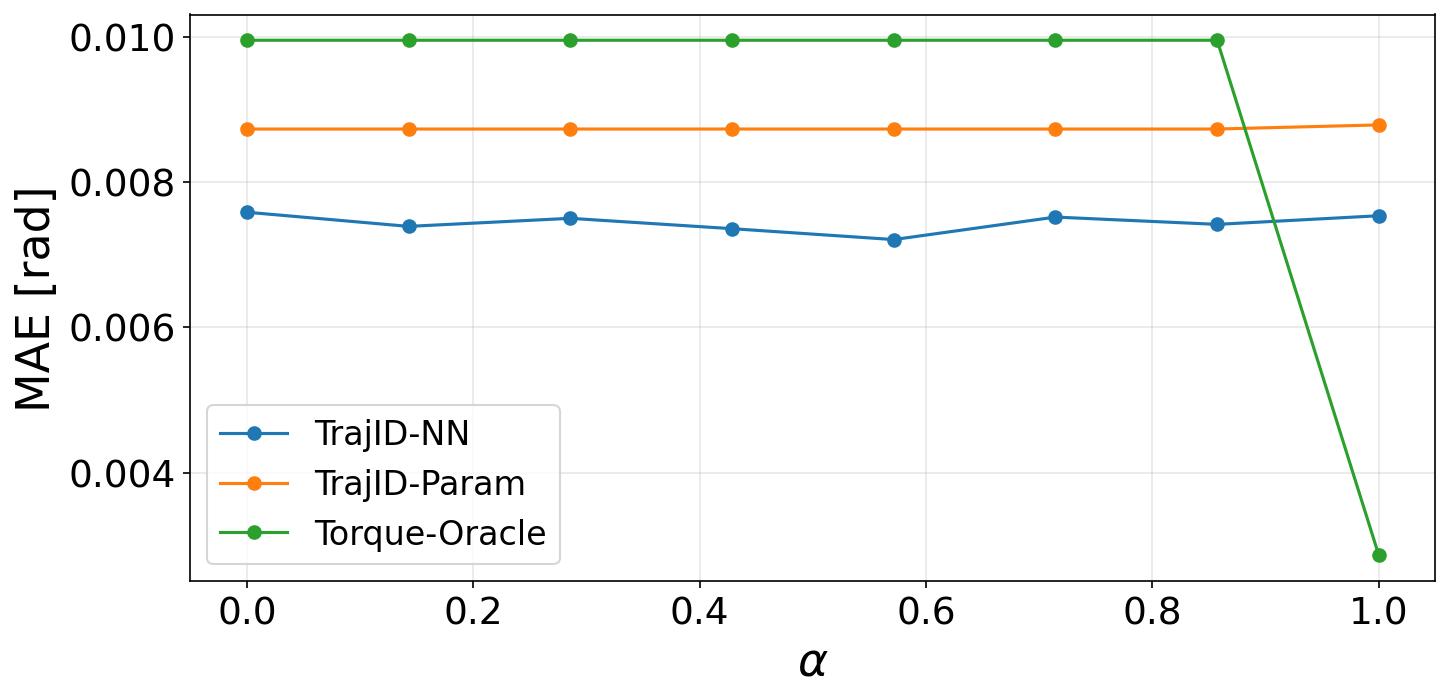}}{\fbox{\parbox{0.9\linewidth}{\centering\small [PLACEHOLDER: experiment2\_alpha\_mae.jpg]}}}
\caption{Validation MAE versus objective weight \(\alpha\) (\(w_q=\alpha\), \(w_{\dot q}=1-\alpha\)) for \textsc{TrajID-Param}, \textsc{TrajID-NN}, and \textsc{Torque-Oracle}.}
\label{fig:w_sensitivity}
\end{figure}

We sweep the objective weights \(W=\mathrm{diag}(w_q, w_{\dot q})\) with \(w_q=\alpha\) and \(w_{\dot q}=1-\alpha\) for \(\alpha\in[0,1]\), keeping all other training settings fixed.
Fig.~\ref{fig:w_sensitivity} shows validation MAE versus \(\alpha\) for \textsc{TrajID-Param}, \textsc{TrajID-NN}, and \textsc{Torque-Oracle}.

\textsc{TrajID-Param} exhibits negligible sensitivity: MAE remains near 8.73~mrad across all \(\alpha\), indicating that the parametric model fits the trajectory regardless of the position–velocity weighting.
\textsc{TrajID-NN} shows mild sensitivity, with best MAE \(\approx 7.21\)~mrad at \(\alpha\approx 0.57\) and slightly higher values (up to \(\approx 7.59\)~mrad) at the extremes.
\textsc{Torque-Oracle} is largely flat for \(\alpha<1\) (MAE \(\approx 9.95\)~mrad) but drops to 2.86~mrad at \(\alpha=1\), where the objective weights position only.
The large error at \(\alpha<1\) stems from overfitting on noisy velocity in the data.

In summary, the parametric model is robust to weight choice, while flexible models trade off overfitting on noisy velocity against position-only weighting.

\subsection{Prediction-Horizon Analysis}
\label{subsec:segment_length_ablation}
\begin{figure}[!t]
\centering
\IfFileExists{experiment3_horizon_mae.jpg}{\includegraphics[width=\linewidth]{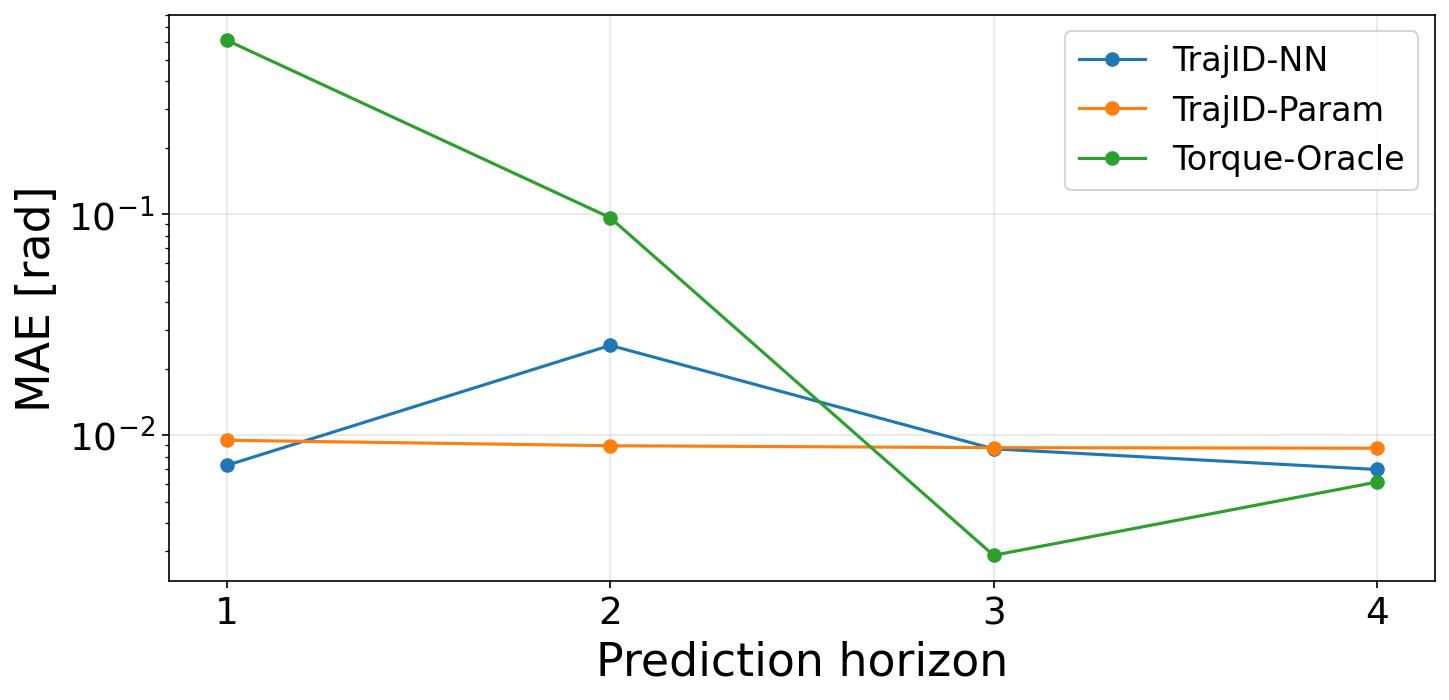}}{\fbox{\parbox{0.9\linewidth}{\centering\small [PLACEHOLDER: experiment3\_horizon\_mae.jpg]}}}
\caption{Validation MAE versus prediction horizon for \textsc{TrajID-Param}, \textsc{TrajID-NN}, and \textsc{Torque-Oracle}.}
\label{fig:segment_length}
\end{figure}

We evaluate validation MAE versus prediction horizon for \textsc{TrajID-Param}, \textsc{TrajID-NN}, and \textsc{Torque-Oracle} under identical training settings.
Fig.~\ref{fig:segment_length} shows the results for horizons 1--4.

\textsc{TrajID-Param} is the most stable across horizons: MAE decreases slightly from 9.49~mrad at horizon 1 to 8.74~mrad at horizon 4, indicating robust long-horizon consistency.
\textsc{TrajID-NN} exhibits a pronounced spike at horizon 2 (25.5~mrad) but recovers to 7.00~mrad at horizon 4, suggesting sensitivity to certain horizon lengths.
\textsc{Torque-Oracle} is highly horizon-dependent: MAE is 612~mrad at horizon 1, drops to 2.86~mrad at horizon 3, then rises to 6.14~mrad at horizon 4.
The poor performance at short horizons reflects the oracle's per-timestep optimization, which can overfit to local noise when the rollout is too short.
Parametric models are nearly insensitive to horizon, unlike the more flexible \textsc{TrajID-NN} and \textsc{Torque-Oracle}.
Models with more parameters tend to overfit on noisy velocity and position data.
Conversely, longer horizons accumulate prediction errors as integration and model mismatch compound over time.

In summary, parametric models offer the most stable horizon behavior. The trade-off between overfitting on noise and error accumulation favors compact parameterizations at moderate horizons.

\section{Scope and Applications}
\label{sec:scope_applications}
    In this section we evaluate how the proposed actuator-identification approach transfers to a real‑world task within the same actuator class considered in the main identification experiments.
    For the miniPi robot~\cite{high2025torque}, we use a structured parameterization focused on MuJoCo internal parameters: \texttt{armature}, \texttt{frictionloss}, and \texttt{damping}, with PD gains held at their default values. We identify these parameters using the same trajectory-matching procedure described in Section~\ref{sec:method} and Appendix~\ref{sec:our_apprach_appendix}, and then use the refined model to train a locomotion policy. The identified MuJoCo parameters for miniPi converged to \texttt{armature} \(=0.0066\,\mathrm{kg\,m^2}\), \texttt{frictionloss} \(=0.1821\), and \texttt{damping} \(=0.2824\,\mathrm{N\,m\,s/rad}\). We interpret these values as physically plausible effective joint-space parameters for the simulator model, rather than as one-to-one estimates of manufacturer-reported hardware quantities, because they are simulator-side modeling terms and the manufacturer documentation does not provide directly comparable inertial or friction parameters for this actuator.
    We cannot include a \textsc{Bench-Sup} locomotion baseline for miniPi because the available test-stand measurements did not provide sufficiently reliable torque labels to train a supervised actuator model suitable for policy learning. We therefore compare only the default (unadjusted) actuator model against the trajectory-identified model in the downstream locomotion study.
    This mirrors common deployment settings in which robots arrive with fixed low‑level controllers and only a default actuator model is available.
    Concretely, in many simulator robot models these internal parameters are not specified and thus remain at simulator defaults; we use this unmodified setting as the baseline to reflect standard RL practice when such parameters are unavailable, rather than a hand-tuned “best possible” physics model.
    We test whether a policy trained on the refined (identified) model outperforms one trained on the default baseline in a forward‑walking task.

    \paragraph{Task.}
    We train two policies in simulation—one using the trajectory-identified actuator model and one using the default actuator model (the default torque-source/PD actuation model, with \texttt{armature}, \texttt{frictionloss}, and \texttt{damping} left at their model-default values)—and compare their real‑robot performance.

    \paragraph{Setup.}
    Training uses LeggedGym~\cite{rudin2022learning} without domain randomization (rewards and details in Appendix~\ref{sec:rl_formulation}, Table~\ref{tab:rl_rewards}). The reward function and all training hyperparameters are held fixed across both policies; only the underlying actuator model differs.
    After training, both policies are deployed on the same miniPi robot under identical conditions.
    To account for inertial measurement unit velocity scaling due to noise, the angular-velocity command on hardware is set to half the simulation value.
    The commanded forward (linear) velocity is set to its maximum to encourage distance covered.
    
    Each policy is evaluated over 10 runs of 7\,s.
    We report mean distance traveled in the commanded direction and mean yaw deviation from the initial heading.

\begin{figure}[!t]
            \centering
            \includegraphics[width=3in]{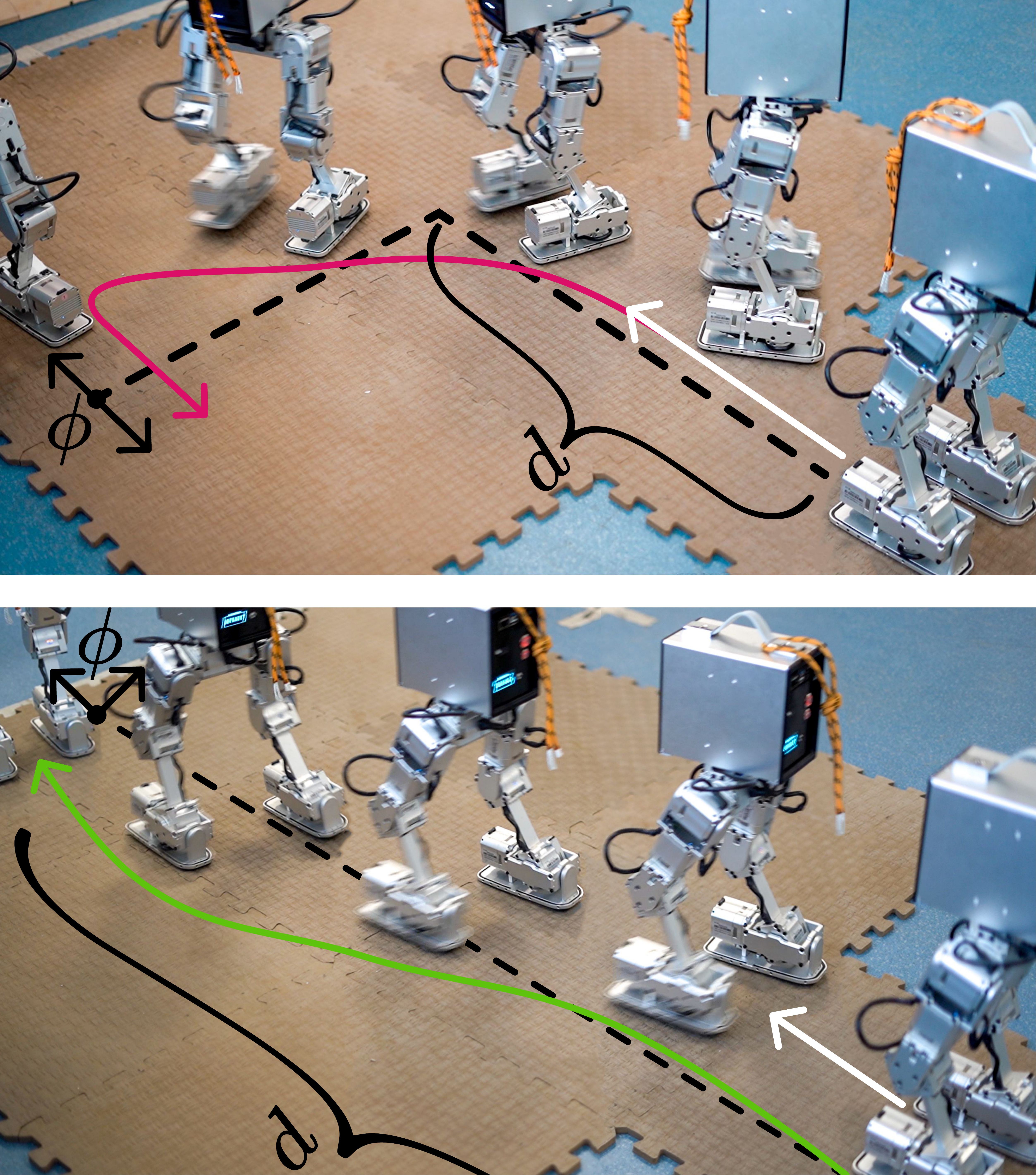}
            \caption{\textbf{Robot Performance Comparison: Refined vs. Baseline RL Policies.}
            The traveled distance is denoted as $d$, and the rotation deviation from the original direction is represented as $\phi$.
            The top image shows robot performance using the baseline model, exhibiting excessive rotation and reduced travel distance.
            The bottom image shows the robot's performance using the refined model, demonstrating improvements in both travel distance and rotational alignment.}
            \label{fig:rl_comparison}
    \end{figure}

    \paragraph{Results.}
    As shown in Fig.~\ref{fig:rl_comparison} and Table~\ref{tab:results}, the baseline policy traveled \(1.12 \pm 0.13\)\,m with \(109 \pm 14.3^\circ\) of rotational deviation, whereas the policy trained on the refined model traveled \(1.64 \pm 0.08\)\,m with only \(27 \pm 6.4^\circ\).
    This corresponds to a \(46\%\) increase in distance and a \(75\%\) reduction in rotation.

    We report these results as supporting downstream validation that the trajectory-identified actuator model offers a consistent practical advantage over the default simulator setting in this locomotion task.

    \begin{table}[tbh]
        \vspace{0.2cm}
            \centering
            \caption{Comparison of our method vs. baseline}
            \begin{tabular}{lcc}
                    \toprule
                    \textbf{Metric} & \textbf{Default Model} & \textbf{TrajID Model} \\
                    & \textbf{(mean \(\pm\) std)} & \textbf{(mean \(\pm\) std)} \\
                    \midrule
                    \multicolumn{3}{c}{\textbf{RL Policy Comparison}} \\
                    Rotation deviation (deg) ↓ & \(109 \pm 14.3\)  & \(27 \pm 6.4\)  \\
                    Travel distance (m) ↑ & \(1.12 \pm 0.13\)  & \(1.64 \pm 0.08\)  \\
                    \bottomrule
            \end{tabular}
            \label{tab:results}
    \end{table}

\section{Conclusion}
\label{sec:conclusion}
We address the \emph{actuation gap} in sim‑to‑real transfer by introducing a practical identification pipeline that learns an effective actuator model directly from observed motion trajectories using a differentiable simulator (MJX).
We formulate identification as trajectory matching and solve it with gradient‑based optimization over actuator and simulator parameters, enabling system‑level calibration without torque sensors or access to internal electrical states.

We validate the approach on a hand‑tuned, high‑gear‑ratio actuator.
Across held‑out rollouts under identical commands, trajectory‑identified actuator models outperformed a stand‑based baseline.
MAE decreased by \(\approx1.88\times\) (from 14.20~mrad for the baseline to 7.54--8.73~mrad with trajectory identification),
indicating that fitting to system‑level trajectories better captures the effective closed‑loop actuation behavior required for high‑fidelity simulation.
Although the framework is model-agnostic in formulation, the present experimental evidence supports conclusions only for the tested actuator class with embedded PD control.

Beyond single‑actuator calibration, improved actuator fidelity translates to downstream control gains.
In a real‑robot locomotion case study (miniPi), a policy trained with the refined motor model showed a consistent practical advantage over the default simulator setting, increasing mean real‑world travel distance from \(1.12 \pm 0.13\)~m to \(1.64 \pm 0.08\)~m and reducing mean rotation deviation from \(109 \pm 14.3^\circ\) to \(27 \pm 6.4^\circ\) under the same protocol.
These results support differentiable trajectory‑level identification as a scalable alternative to purely steady‑state test-stand characterization when accurate closed‑loop motion reproduction is the goal.

We summarize key advantages and limitations below.
\paragraph{Advantages}
\begin{itemize}
    \item \textbf{System-level identification:} learns an \emph{effective} actuator model directly from closed-loop trajectories, targeting the behavior that matters for simulation fidelity.
    \item \textbf{Torque sensor-free:} requires only commanded inputs and encoder motion; no torque sensors or electrical measurements are needed for identification.
    \item \textbf{Quantitative accuracy gains:} improves held-out trajectory alignment versus a stand-based steady-state baseline, reducing MAE from 14.20~mrad to as low as 7.54~mrad (\(\approx\)1.88\(\times\)).
    \item \textbf{Downstream impact:} improves real-robot locomotion performance, increasing travel distance by 46\% and reducing rotational deviation by 75\% under the same protocol.
\end{itemize}

\paragraph{Limitations}
\begin{itemize}
    \item Model expressiveness is bounded by the chosen parameterization; unmodeled effects (e.g., temperature drift, battery‑voltage sag, hysteresis, contact mismatch) can remain as residual errors.
    \item Performance depends on excitation richness and sensor quality; long‑horizon rollouts can accumulate integration error.
    \item Results are reported for a single actuator class and a limited number of hardware instances; broader validation across actuator technologies and platforms is left as future work.
\end{itemize}

Several directions remain open for future work.
The parameterization can be broadened to capture additional non‑idealities such as dead zones, rate limits, and explicit delay.
Robustness to noise and non‑stationarity can be improved through regularization and dataset design, including systematic analysis of \textsc{Torque-Oracle} overfitting.
Scaling to multi‑joint robots and contact‑rich tasks with standardized sim‑to‑real benchmarks is a natural next step.
Recurrent and history-based neural architectures offer a path toward capturing hysteresis, delay, and other path-dependent effects that instantaneous models cannot represent, and may help reduce the remaining gap between \textsc{TrajID-NN} and \textsc{Torque-Oracle}.
Finally, broader RL ablations would help isolate the actuator-model contribution to downstream locomotion performance.

\appendices
\section{\break Motor test stand: dataset characterization and actuator-model learning}
\label{app:bench}

The actuator under test \cite{starkit_servo_alum} has a high gear ratio (210:1), a 72\,MHz microcontroller, and operates from a 12\,V supply.
The embedded low‑level controller is a hand‑tuned PD regulator.
We adopt a simple outer‑loop structure that maps desired position and measured motion to a pulse-width modulation (PWM) duty cycle, which modulates phase current and thereby torque.
The resulting control pathway is \(q^{\mathrm{des}},\, q,\, \dot q \rightarrow u \rightarrow I \rightarrow \tau\), where \(I\) denotes the phase current.
Here \(u\in[-1,1]\) denotes the PWM duty cycle, computed as

\begin{equation}
u = \mathrm{clip}\!\left(K_p (q^{\mathrm{des}}-q) - K_d \dot{q},\, -1,\, 1\right).
\label{eq:pwm_control}
\end{equation}
 
This controller implements a PD mapping from position error and velocity to the PWM duty cycle; the gains \(K_p\) and \(K_d\) are dimensionless and tuned empirically for stable tracking.
The PWM duty cycle drives the motor via the power stage, effectively shaping the average phase current and the output torque.

The mapping from current to torque is treated as a black box that aggregates electrical and mechanical effects (e.g., back electromotive force, current limits, drivetrain friction, and gear‑train inertia).

\begin{figure}[t]
\centering
\includegraphics[width=\linewidth]{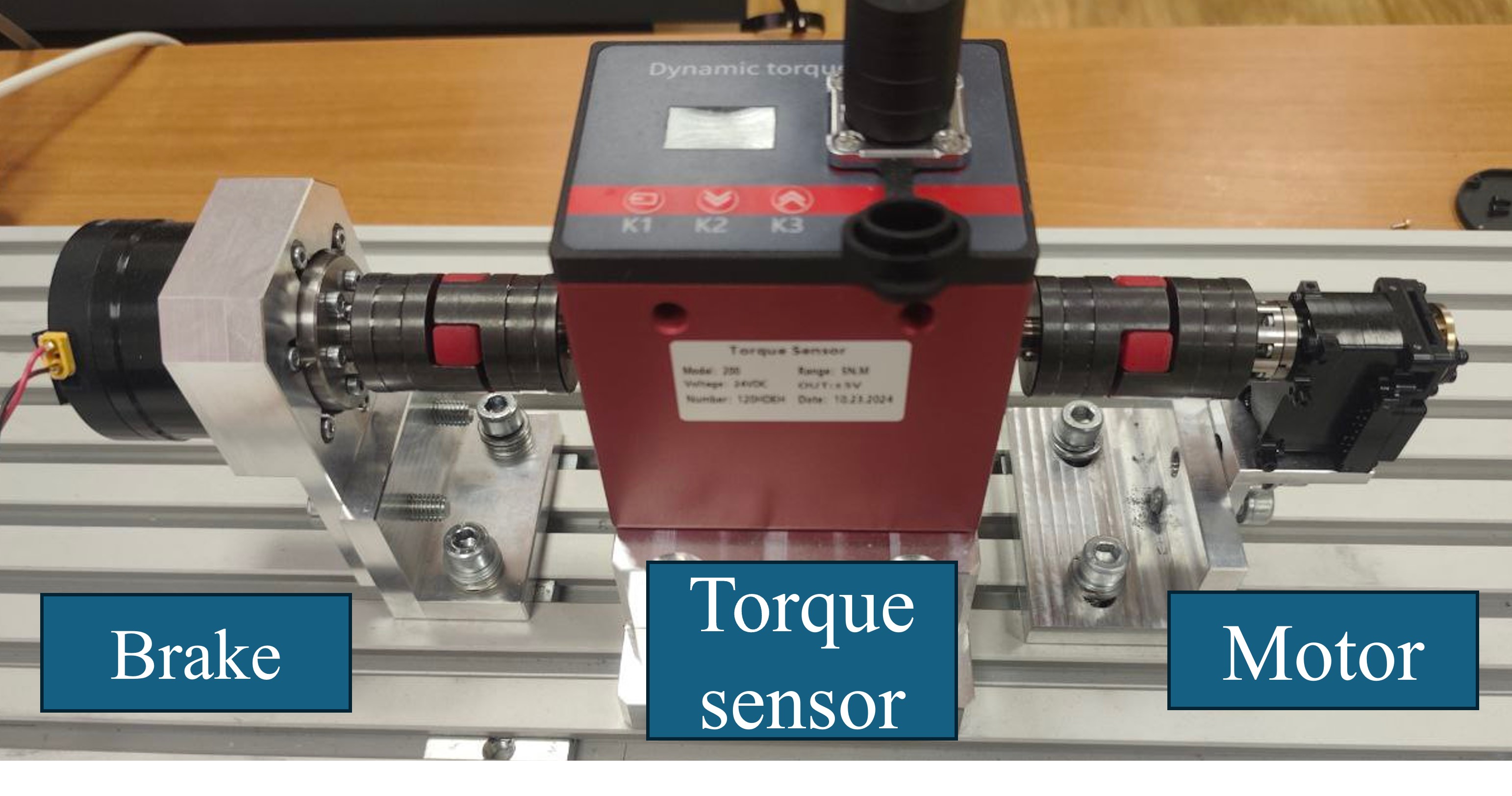}
\caption{\textbf{Motor test-stand setup:} Experimental apparatus used to characterize the actuator under controlled loading. The rig couples a test motor and a load (brake) motor through an inline torque transducer, enabling synchronized torque/velocity measurements.}
\label{fig:motor_setup}
\end{figure}

The motor is characterized on a dedicated test stand (Fig.~\ref{fig:motor_setup}) comprising a driven motor, a brake motor, and a DYN-200 inline torque sensor mounted on a rigid bench and coupled on a common shaft.
To avoid uncertainty from unknown composite inertia, we restrict measurements to steady-state operating points: we command a PWM duty cycle \(u\) while the brake motor applies increasing resistive torque, and we record the resulting steady angular velocity \(\dot q\) and output torque \(\tau\).
These data define the steady-state map \((u,\dot q)\!\rightarrow\!\tau\).
Non-steady segments with high acceleration are filtered out prior to analysis.

Using the steady‑state dataset, we train a compact feed‑forward MLP to predict torque \(\tau\) from \((u,\dot q)\).
The MLP uses two rectified linear unit hidden layers with widths 128 and 64, followed by a 1‑D torque output (Linear \(2\!\rightarrow\!128\!\rightarrow\!64\!\rightarrow\!1\)); inputs are standardized (mean/scale) and the network has 8,705 trainable parameters.
For trajectory‑level experiments, the PD law above maps \((q^{\mathrm{des}}, q, \dot q)\) to \(u\), after which the learned map outputs \(\tau\), yielding the overall mapping \((q^{\mathrm{des}}, q, \dot q)\!\rightarrow\!\tau\).

\begin{figure}[t]
\centering
\includegraphics[width=\linewidth]{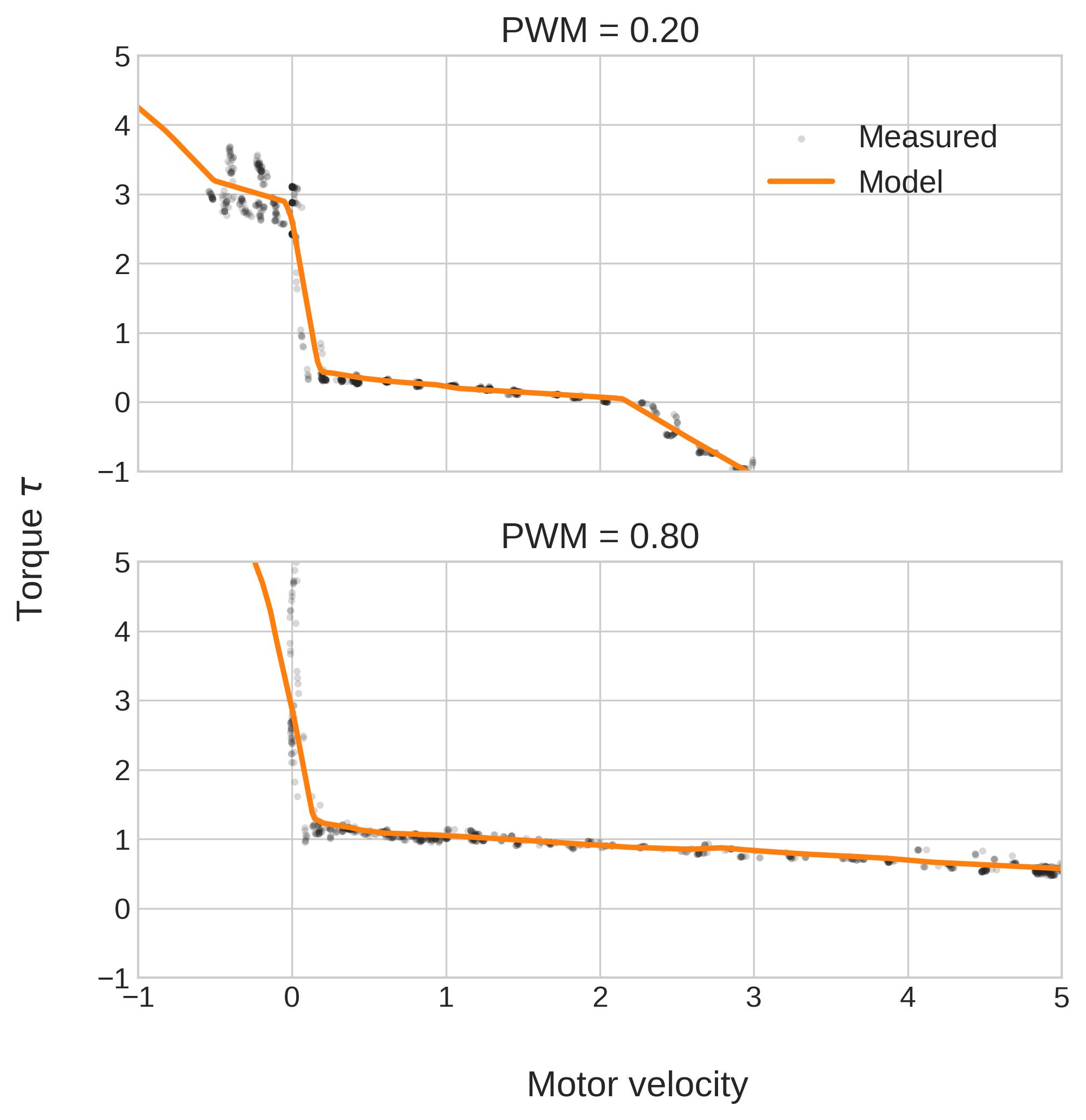}
\caption{\textbf{Measured vs.\ model-predicted torque.} Steady-state motor torque \(\tau\) (N\,m) versus motor velocity (rad/s) for two PWM duty-cycle commands (\(u=0.2\) and \(u=0.8\)). Gray markers show downsampled test-stand measurements; the orange curve shows the learned (\textsc{Bench-Sup}) model evaluated on a dense velocity grid.}
\label{fig:pwm_model_vs_measured}
\end{figure}

Fig.~\ref{fig:pwm_model_vs_measured} compares the learned map with measured steady‑state torque data.
The relationship is approximately linear when the commanded direction and measured velocity agree, while a pronounced near‑zero‑velocity region exhibits increased resistive torque (braking behavior). During braking tests, we observed repeated shaft damage when the braking torque exceeded approximately \(7\)-\(8\)~N\,m. To avoid further hardware failure, we limited subsequent data collection to approximately \(5\)-\(6\)~N\,m. As a result, the \textsc{Bench-Sup} dataset under-samples the high-load, low-speed braking regime near zero velocity. We therefore clarify that \textsc{Bench-Sup} is less well supported in this operating region than in the rest of the test-stand map, and that this coverage gap likely biases the comparison against \textsc{Bench-Sup} in this regime. The resulting test-stand map serves as a practical baseline for the trajectory‑level comparisons in the main text.

\section{\break Trajectory dataset and identification setup}
\label{sec:our_apprach_appendix}

\subsection{Overview}
This appendix details how we collect and preprocess the \emph{trajectory dataset} used to identify \textsc{TrajID-Param}, \textsc{TrajID-NN}, \textsc{Residual-RL}, and \textsc{Torque-Oracle}. The \textsc{Bench-Sup} baseline is trained separately from dedicated test-stand measurements (see Appendix~\ref{app:bench}).

\begin{figure}[t]
\centering
\includegraphics[width=0.5\linewidth]{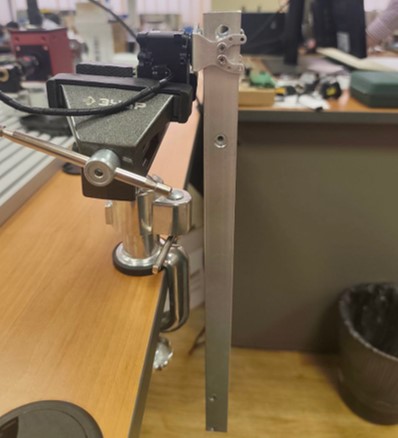}
\caption{Table-top setup with actuator and rigid rod load used to collect identification trajectories.}
\label{fig:traj_rig}
\end{figure}

\paragraph{Experimental setup.}
The actuator is mounted on a table-top rig with a known external load to induce repeatable, nontrivial dynamics (Fig.~\ref{fig:traj_rig}).
In our setup, the load is a rigid bar attached to the motor shaft, with mass \(m=0.24\,\mathrm{kg}\) and dimensions \(0.04\times0.02\times0.352\,\mathrm{m}\) (width \(\times\) thickness \(\times\) length).

\paragraph{Desired trajectory generation.}
The optimization in Eq.~\eqref{eq:final_opt_problem} requires real trajectories \(\{s_i\}\) under known commands \(\{a_i\}\).
The measured state is

\begin{equation}
s_i = [\,q_i,\dot{q}_i\,]^T
\label{eq:state_definition}
\end{equation}

where \(q_i\) and \(\dot{q}_i\) are joint angle and velocity at timestep \(i\).

We synthesize commanded velocity profiles as a sum of randomly sampled Fourier modes to ensure diverse excitation:

\begin{equation}
\dot{q}^{\mathrm{des}}(t) = \sum_{k=1}^{K} A_k \sin(\omega_k t + \phi_k).
\label{eq:fourier_velocity}
\end{equation}
 
The number of modes \(K\), amplitudes \(A_k\), frequencies \(\omega_k\), and phases \(\phi_k\) are sampled uniformly from predefined ranges, and \(\dot{q}^{\mathrm{des}}\) is clipped to hardware limits \(|\dot{q}^{\mathrm{des}}(t)| \le \dot{q}_{\max}\).
\(\dot{q}^{\mathrm{des}}(t)\) is only an intermediate signal used to shape frequency content and amplitude.
We chose this random-Fourier design empirically to increase dataset diversity rather than to satisfy a formal persistence-of-excitation condition. In practice, it produces a mixture of transient segments, near-limit operating points where the actuator approaches its velocity bound, and quasi-steady regions, improving coverage of the actuator’s relevant operating regime. We therefore present this excitation as an empirical strategy for improving practical parameter observability, not as a formal identifiability guarantee. We then integrate \(\dot{q}^{\mathrm{des}}\) to obtain the position command \(q^{\mathrm{des}}(t)\).
We set the action to the desired position, \(a_i = q^{\mathrm{des}}_i\).
Applied via the embedded position servo, the hardware executes \(q^{\mathrm{des}},\, q,\, \dot q \rightarrow u \rightarrow I \rightarrow \tau\) (Appendix~\ref{app:bench}).
We treat the command-to-torque stack as a fixed black-box mapping that we seek to identify.

We collect training trajectories of duration \(T_{\mathrm{train}} = 40\,\mathrm{s}\) and one test trajectory of duration \(T_{\mathrm{test}} = 10\,\mathrm{s}\).
All trajectories are logged at \(\delta = 0.002\,\mathrm{s}\) (500\,Hz).

\paragraph{Train/test split, segmentation, and simulator settings.}
During optimization, training trajectories are divided into short segments of length \(N=3\) (i.e., \(N\delta = 6\,\mathrm{ms}\)) for the main trajectory-matching results. We choose \(N=3\) as a compromise between short segments (which tend to overfit on noise) and long segments (which accumulate prediction errors), as discussed in Section~\ref{subsec:segment_length_ablation}. Segments are sampled with overlap.
For \(\Phi_z(s,a,\delta)\), we use MuJoCo's explicit Euler integrator for efficiency.
In Eq.~\eqref{eq:final_opt_problem}, we set \(W=\mathrm{diag}(1,0)\), i.e., \(w_q=1\), \(w_{\dot q}=0\), so the objective penalizes position error only. Velocity remains part of the observed state and simulator rollout, but velocity residuals are not penalized because the measured velocity signal is noticeably noisier than position. The sensitivity sweep over \(W=\mathrm{diag}(\alpha,1-\alpha)\) in Section~\ref{subsec:w_sensitivity} shows that flexible models overfit on velocity when it is weighted, whereas position-only weighting yields robust results across models.

All identified models minimize the segmented trajectory-matching objective in Eq.~\eqref{eq:final_opt_problem} using Adam (Optax) with learning rate \(10^{-2}\). In our experiments, \(M = 2{,}000\) sampled trajectory segments are used in each optimization step.
Early stopping uses patience \(=200\) and \(\mathrm{min\_delta}=0\).
We retain the parameters with the lowest loss on the entire training dataset.

\paragraph{Neural actuator model (\textsc{TrajID-NN}).}
The neural actuator model is a feed‑forward MLP that maps the concatenated input \([q^{\mathrm{des}},q,\dot q]\in\mathbb{R}^3\) to torque \(\tau\in\mathbb{R}\), with architecture \([3,32,32,1]\) (two hidden layers of width 32; 1{,}217 trainable parameters).
No recurrence or history is used.
Inputs are normalized to zero mean and unit variance.
Training follows Eq.~\eqref{eq:final_opt_problem} with the same segmentation and optimizer as above.

\paragraph{ASAP-inspired residual baseline (\textsc{Residual-RL}).}
To compare against action-space compensation methods, we add \textsc{Residual-RL}, an ASAP-inspired proxy that learns a residual position correction \(\delta a_i\) instead of simulator parameters or torque outputs.
At timestep \(i\), the model observes the current rollout state and produces a corrected command
\begin{equation}
\tilde a_i = a_i + \delta a_i.
\end{equation}
The corrected position command is then converted to torque by a fixed PD regulator with gains \(k_p=5\) and \(k_v=1\); these values were chosen near the optimum identified by \textsc{TrajID-Param}.
Unlike the original ASAP method, which is implemented in a different simulator stack and for a different robot/control pipeline, \textsc{Residual-RL} is implemented in the same simulator environment as the other compared models for fairness.
To keep model capacity comparable, the residual policy uses the same hidden-layer widths as \textsc{TrajID-NN}.

\noindent The \textsc{Residual-RL} policy observes \((q_i,\dot q_i,q^{\mathrm{des}}_i)\) at each timestep.

\noindent The residual policy is trained with Stable-Baselines3 PPO~\cite{stable-baselines3}. Its reward is implemented as the negative rollout loss corresponding to Eq.~\eqref{eq:final_opt_problem}, so that PPO maximizes the same objective that the identification models minimize. Reward normalization is enabled, as we observed substantially improved learning with normalized rewards.

\noindent The residual policy uses a feed-forward MLP with dimensions \([3,32,32,1]\), i.e., two hidden layers of width 32 and a scalar residual-action output, matching the hidden-layer widths of \textsc{TrajID-NN}.

\noindent PPO training uses 8 million timesteps with 256 parallel environments and 256 rollout steps per environment, and takes 46.34 minutes in wall-clock time. We use learning rate \(4\times10^{-4}\), minibatch size 64, 10 PPO epochs per update, discount factor \(\gamma=1.0\), GAE parameter \(\lambda=0.95\), clip range 0.2, entropy coefficient 0.0, and value-loss coefficient 0.5.

\section{\break Reinforcement Learning Problem Formulation}
\label{sec:rl_formulation}

For completeness, we provide the RL problem setup used in the experiments.
This includes the state and action definitions, reward functions (Table~\ref{tab:rl_rewards}), and any scaling or normalization applied to the observations or actions.
Specific formulations for locomotion tasks, reward shaping, and termination conditions are detailed here to allow reproducibility and comparison with prior work.

\begin{table}[t]
\centering
\caption{\textbf{Reward terms used in the RL experiments.}}
\label{tab:rl_rewards}
\small
\setlength{\tabcolsep}{5pt}
\renewcommand{\arraystretch}{1.4}
\begin{tabular}{p{0.28\linewidth} >{\centering\arraybackslash}p{0.52\linewidth} p{0.12\linewidth}}
\toprule
\textbf{Term} & \textbf{Formula} & \textbf{Scale} \\
\midrule
Tracking linear velocity & \(\exp\!\big(-\tfrac{(v-v^{\mathrm{cmd}})^2}{\sigma^2}\big)\) for \(v_x,v_y\) & 1.5\\
Tracking angular velocity (yaw) & \(\exp\!\big(-4\,(\omega_z-\omega^{\mathrm{cmd}}_z)^2\big)\) & 1.5\\
\addlinespace[4pt]
Base upright (roll/pitch) & \(\exp\!\big(-30\,\lVert[\mathrm{roll},\mathrm{pitch}]\rVert_2\big)\) & 0.9\\
Base height & \(\exp\!\big(-500\,(z/z_{\mathrm{target}}-1)^2\big)\) & 0.8\\
\addlinespace[4pt]
Base acceleration penalty & \(\exp\!\big(-\sum_j |a^{\mathrm{root}}_j|\big)\) & 0.1\\
Action smoothness & \(\lVert a_t-a_{t-1}\rVert^2 + \lVert a_t + a_{t-2}-2a_{t-1}\rVert^2\) & -0.02\\
\addlinespace[4pt]
Joint tracking to reference & \(\exp\!\big(-6\,\lVert q-q_{\mathrm{ref}}\rVert_2\big)\) & 1.5\\
Default posture & \(\exp\!\big(-\lVert q-q_{\mathrm{def}}\rVert_2\big)\) & 0.5\\
\addlinespace[4pt]
Feet clearance & Reward when foot height near target during swing & 0.8\\
Feet orientation & \(\exp\!\big(-(\sum|\mathrm{roll}|+|\mathrm{pitch}|)\big)\) on feet & 0.3\\
Ankle roll regularization & \(\exp\!\big(-\sum|\theta_{\mathrm{ankle\_roll}}|\big)\) & 2.0\\
\bottomrule
\end{tabular}
\end{table}

\paragraph{Task interface and parameters.}
Observations (48 dims): base angular velocity, gravity, velocity commands \([v_x,v_y,\omega_z]\), standing mask, gait phase, joint positions and velocities, last action. In the velocity-tracking reward, \(\sigma\) is the acceptance bandwidth (speed tolerance).
Actions: target joint positions for 12 actuators, scaled as \(q^{\mathrm{target}} = q^{\mathrm{def}} + 0.25\,a\).
Simulation: timestep 1\,ms, policy at 100\,Hz; PhysX.
Episode length 16\,s. Command ranges: \(v_x\in[-0.4,0.4]\) m/s, \(v_y\in[-0.25,0.25]\) m/s, \(\omega_z\in[-0.6,0.6]\) rad/s.
Control: position control with joint stiffness and damping (hip, thigh, calf, ankle).
Gait: cycle time 0.4\,s, target foot height 0.02\,m, base height 0.345\,m.
Termination: on timeout or base contact force \(>1\) N.
Training (Proximal Policy Optimization): Kullback–Leibler divergence target 0.01, 3001 max updates.

\bibliographystyle{IEEEtran}
\bibliography{references}

\begin{thebibliography}{10}
\providecommand{\url}[1]{#1}
\csname url@samestyle\endcsname
\providecommand{\newblock}{\relax}
\providecommand{\bibinfo}[2]{#2}
\providecommand{\BIBentrySTDinterwordspacing}{\spaceskip=0pt\relax}
\providecommand{\BIBentryALTinterwordstretchfactor}{4}
\providecommand{\BIBentryALTinterwordspacing}{\spaceskip=\fontdimen2\font plus
\BIBentryALTinterwordstretchfactor\fontdimen3\font minus \fontdimen4\font\relax}
\providecommand{\BIBforeignlanguage}[2]{{%
\expandafter\ifx\csname l@#1\endcsname\relax
\typeout{** WARNING: IEEEtran.bst: No hyphenation pattern has been}%
\typeout{** loaded for the language `#1'. Using the pattern for}%
\typeout{** the default language instead.}%
\else
\language=\csname l@#1\endcsname
\fi
#2}}
\providecommand{\BIBdecl}{\relax}
\BIBdecl

\bibitem{di2018dynamic}
J.~Di~Carlo, P.~M. Wensing, B.~Katz, G.~Bledt, and S.~Kim, ``Dynamic locomotion in the mit cheetah 3 through convex model-predictive control,'' in \emph{2018 IEEE/RSJ international conference on intelligent robots and systems (IROS)}.\hskip 1em plus 0.5em minus 0.4em\relax IEEE, 2018, pp. 1--9.

\bibitem{kim2019highly}
D.~Kim, J.~Di~Carlo, B.~Katz, G.~Bledt, and S.~Kim, ``Highly dynamic quadruped locomotion via whole-body impulse control and model predictive control,'' \emph{arXiv preprint arXiv:1909.06586}, 2019.

\bibitem{gaertner2021collision}
M.~Gaertner, M.~Bjelonic, F.~Farshidian, and M.~Hutter, ``Collision-free mpc for legged robots in static and dynamic scenes,'' in \emph{2021 IEEE International Conference on Robotics and Automation (ICRA)}.\hskip 1em plus 0.5em minus 0.4em\relax IEEE, 2021, pp. 8266--8272.

\bibitem{hwangbo2019learning}
J.~Hwangbo, J.~Lee, A.~Dosovitskiy, D.~Bellicoso, V.~Tsounis, V.~Koltun, and M.~Hutter, ``Learning agile and dynamic motor skills for legged robots,'' \emph{Science Robotics}, vol.~4, no.~26, p. eaau5872, 2019.

\bibitem{james2019sim}
S.~James, P.~Wohlhart, M.~Kalakrishnan, D.~Kalashnikov, A.~Irpan, J.~Ibarz, S.~Levine, R.~Hadsell, and K.~Bousmalis, ``Sim-to-real via sim-to-sim: Data-efficient robotic grasping via randomized-to-canonical adaptation networks,'' in \emph{Proceedings of the IEEE/CVF conference on computer vision and pattern recognition}, 2019, pp. 12\,627--12\,637.

\bibitem{aljalbout2025reality}
E.~Aljalbout, J.~Xing, A.~Romero, I.~Akinola, C.~R. Garrett, E.~Heiden, A.~Gupta, T.~Hermans, Y.~Narang, D.~Fox \emph{et~al.}, ``The reality gap in robotics: Challenges, solutions, and best practices,'' \emph{Annual Review of Control, Robotics, and Autonomous Systems}, vol.~9, 2025.

\bibitem{xie2023learning}
Z.~Xie, P.~Gergondet, F.~Kanehiro \emph{et~al.}, ``Learning bipedal walking for humanoids with current feedback,'' \emph{IEEE Access}, vol.~11, pp. 82\,013--82\,023, 2023.

\bibitem{siekmann2021sim}
J.~Siekmann, Y.~Godse, A.~Fern, and J.~Hurst, ``Sim-to-real learning of all common bipedal gaits via periodic reward composition,'' in \emph{2021 IEEE International Conference on Robotics and Automation (ICRA)}.\hskip 1em plus 0.5em minus 0.4em\relax IEEE, 2021, pp. 7309--7315.

\bibitem{rodriguez2021deepwalk}
D.~Rodriguez and S.~Behnke, ``Deepwalk: Omnidirectional bipedal gait by deep reinforcement learning,'' in \emph{2021 IEEE international conference on robotics and automation (ICRA)}.\hskip 1em plus 0.5em minus 0.4em\relax IEEE, 2021, pp. 3033--3039.

\bibitem{li2024reinforcement}
Z.~Li, X.~B. Peng, P.~Abbeel, S.~Levine, G.~Berseth, and K.~Sreenath, ``Reinforcement learning for versatile, dynamic, and robust bipedal locomotion control,'' \emph{The International Journal of Robotics Research}, p. 02783649241285161, 2024.

\bibitem{ravichandar2025preferenced}
P.~Ravichandar, L.~Krishna, N.~Sobanbabu, and Q.~Nguyen, ``Preferenced oracle guided multi-mode policies for dynamic bipedal loco-manipulation,'' in \emph{2025 IEEE/RSJ International Conference on Intelligent Robots and Systems (IROS)}.\hskip 1em plus 0.5em minus 0.4em\relax IEEE, 2025, pp. 6600--6606.

\bibitem{zhang2025natural}
H.~Zhang, L.~Zhang, Z.~Chen, L.~Chen, Y.~Wang, and R.~Xiong, ``Natural humanoid robot locomotion with generative motion prior,'' \emph{arXiv preprint arXiv:2503.09015}, 2025.

\bibitem{rudin2022learning}
N.~Rudin, D.~Hoeller, P.~Reist, and M.~Hutter, ``Learning to walk in minutes using massively parallel deep reinforcement learning,'' in \emph{Conference on Robot Learning}.\hskip 1em plus 0.5em minus 0.4em\relax PMLR, 2022, pp. 91--100.

\bibitem{makoviychuk2021isaac}
V.~Makoviychuk, L.~Wawrzyniak, Y.~Guo, M.~Lu, K.~Storey, M.~Macklin, D.~Hoeller, N.~Rudin, A.~Allshire, A.~Handa \emph{et~al.}, ``Isaac gym: High performance gpu-based physics simulation for robot learning,'' \emph{arXiv preprint arXiv:2108.10470}, 2021.

\bibitem{mittal2025isaaclab}
\BIBentryALTinterwordspacing
M.~Mittal, P.~Roth, J.~Tigue, A.~Richard, O.~Zhang, P.~Du, A.~Serrano-Muñoz, X.~Yao, R.~Zurbrügg, N.~Rudin, L.~Wawrzyniak, M.~Rakhsha, A.~Denzler, E.~Heiden, A.~Borovicka, O.~Ahmed, I.~Akinola, A.~Anwar, M.~T. Carlson, J.~Y. Feng, A.~Garg, R.~Gasoto, L.~Gulich, Y.~Guo, M.~Gussert, A.~Hansen, M.~Kulkarni, C.~Li, W.~Liu, V.~Makoviychuk, G.~Malczyk, H.~Mazhar, M.~Moghani, A.~Murali, M.~Noseworthy, A.~Poddubny, N.~Ratliff, W.~Rehberg, C.~Schwarke, R.~Singh, J.~L. Smith, B.~Tang, R.~Thaker, M.~Trepte, K.~V. Wyk, F.~Yu, A.~Millane, V.~Ramasamy, R.~Steiner, S.~Subramanian, C.~Volk, C.~Chen, N.~Jawale, A.~V. Kuruttukulam, M.~A. Lin, A.~Mandlekar, K.~Patzwaldt, J.~Welsh, H.~Zhao, F.~Anes, J.-F. Lafleche, N.~Moënne-Loccoz, S.~Park, R.~Stepinski, D.~V. Gelder, C.~Amevor, J.~Carius, J.~Chang, A.~H. Chen, P.~de~Heras~Ciechomski, G.~Daviet, M.~Mohajerani, J.~von Muralt, V.~Reutskyy, M.~Sauter, S.~Schirm, E.~L. Shi, P.~Terdiman, K.~Vilella, T.~Widmer, G.~Yeoman, T.~Chen, S.~Grizan, C.~Li, L.~Li, C.~Smith, R.~Wiltz, K.~Alexis, Y.~Chang, D.~Chu, L.~J. Fan, F.~Farshidian, A.~Handa, S.~Huang, M.~Hutter, Y.~Narang, S.~Pouya, S.~Sheng, Y.~Zhu, M.~Macklin, A.~Moravanszky, P.~Reist, Y.~Guo, D.~Hoeller, and G.~State, ``Isaac lab: A gpu-accelerated simulation framework for multi-modal robot learning,'' \emph{arXiv preprint arXiv:2511.04831}, 2025. [Online]. Available: \url{https://arxiv.org/abs/2511.04831}
\BIBentrySTDinterwordspacing

\bibitem{schmidt2022practical}
A.~Schmidt, T.~Gumpert, S.~Schreiber, and A.~Albu-Sch{\"a}ffer, ``Practical approach to characterize realistic motor dynamics for robotic simulation independent of the use case,'' in \emph{2022 IEEE/ASME International Conference on Advanced Intelligent Mechatronics (AIM)}.\hskip 1em plus 0.5em minus 0.4em\relax IEEE, 2022, pp. 1144--1151.

\bibitem{bittencourt2010extended}
A.~C. Bittencourt, E.~Wernholt, S.~Sander-Tavallaey, and T.~Brog{\aa}rdh, ``An extended friction model to capture load and temperature effects in robot joints,'' in \emph{2010 IEEE/RSJ international conference on intelligent robots and systems}.\hskip 1em plus 0.5em minus 0.4em\relax IEEE, 2010, pp. 6161--6167.

\bibitem{wolf2018extending}
S.~Wolf and M.~Iskandar, ``Extending a dynamic friction model with nonlinear viscous and thermal dependency for a motor and harmonic drive gear,'' in \emph{2018 IEEE International Conference on Robotics and Automation (ICRA)}.\hskip 1em plus 0.5em minus 0.4em\relax IEEE, 2018, pp. 783--790.

\bibitem{wang2018advanced}
F.~Wang, Z.~Zhang, X.~Mei, J.~Rodr{\'\i}guez, and R.~Kennel, ``Advanced control strategies of induction machine: Field oriented control, direct torque control and model predictive control,'' \emph{energies}, vol.~11, no.~1, p. 120, 2018.

\bibitem{zhang2016overview}
Y.~Zhang, B.~Xia, H.~Yang, and J.~Rodriguez, ``Overview of model predictive control for induction motor drives,'' \emph{Chinese Journal of Electrical Engineering}, vol.~2, no.~1, pp. 62--76, 2016.

\bibitem{MARTYR2007144}
\BIBentryALTinterwordspacing
A.~Martyr and M.~Plint, ``8 - dynamometers and the measurement of torque,'' in \emph{Engine Testing (Third Edition)}, third edition~ed., A.~Martyr and M.~Plint, Eds.\hskip 1em plus 0.5em minus 0.4em\relax Oxford: Butterworth-Heinemann, 2007, pp. 144--169. [Online]. Available: \url{https://www.sciencedirect.com/science/article/pii/B9780750684392500116}
\BIBentrySTDinterwordspacing

\bibitem{sziki2022measurement}
G.~Sziki, A.~Szanto, J.~Kiss, G.~Juhasz, and E.~Adamko, ``Measurement system for the experimental study and testing of electric motors at the faculty of engineering,'' \emph{University of Debrecen. Applied Sciences, 12 (19)}, pp. 1--18, 2022.

\bibitem{lee2023performance}
T.-W. Lee and D.-K. Hong, ``Performance validation of high-speed motor for electric turbochargers using various test methods,'' \emph{Electronics}, vol.~12, no.~13, p. 2937, 2023.

\bibitem{akkaya2019solving}
I.~Akkaya, M.~Andrychowicz, M.~Chociej, M.~Litwin, B.~McGrew, A.~Petron, A.~Paino, M.~Plappert, G.~Powell, R.~Ribas \emph{et~al.}, ``Solving rubik's cube with a robot hand,'' \emph{arXiv preprint arXiv:1910.07113}, 2019.

\bibitem{tiboni2023domain}
G.~Tiboni, P.~Klink, J.~Peters, T.~Tommasi, C.~D'Eramo, and G.~Chalvatzaki, ``Domain randomization via entropy maximization,'' \emph{arXiv preprint arXiv:2311.01885}, 2023.

\bibitem{haarnoja2024learning}
T.~Haarnoja, B.~Moran, G.~Lever, S.~H. Huang, D.~Tirumala, J.~Humplik, M.~Wulfmeier, S.~Tunyasuvunakool, N.~Y. Siegel, R.~Hafner \emph{et~al.}, ``Learning agile soccer skills for a bipedal robot with deep reinforcement learning,'' \emph{Science Robotics}, vol.~9, no.~89, p. eadi8022, 2024.

\bibitem{duclusaud2025extended}
M.~Duclusaud, G.~Passault, V.~Padois, and O.~Ly, ``Extended friction models for the physics simulation of servo actuators,'' in \emph{2025 IEEE International Conference on Robotics and Automation (ICRA)}.\hskip 1em plus 0.5em minus 0.4em\relax IEEE, 2025, pp. 12\,091--12\,097.

\bibitem{brax2021github}
\BIBentryALTinterwordspacing
C.~D. Freeman, E.~Frey, A.~Raichuk, S.~Girgin, I.~Mordatch, and O.~Bachem, ``Brax - a differentiable physics engine for large scale rigid body simulation,'' 2021. [Online]. Available: \url{http://github.com/google/brax}
\BIBentrySTDinterwordspacing

\bibitem{heiden2021neuralsim}
\BIBentryALTinterwordspacing
E.~Heiden, D.~Millard, E.~Coumans, Y.~Sheng, and G.~S. Sukhatme, ``Neural{S}im: Augmenting differentiable simulators with neural networks,'' in \emph{Proceedings of the IEEE International Conference on Robotics and Automation (ICRA)}, 2021. [Online]. Available: \url{https://github.com/google-research/tiny-differentiable-simulator}
\BIBentrySTDinterwordspacing

\bibitem{erez2015simulation}
T.~Erez, Y.~Tassa, and E.~Todorov, ``Simulation tools for model-based robotics: Comparison of bullet, havok, mujoco, ode and physx,'' in \emph{2015 IEEE international conference on robotics and automation (ICRA)}.\hskip 1em plus 0.5em minus 0.4em\relax IEEE, 2015, pp. 4397--4404.

\bibitem{liao2023performance}
C.~Liao, Y.~Wang, X.~Ding, Y.~Ren, X.~Duan, and J.~He, ``Performance comparison of typical physics engines using robot models with multiple joints,'' \emph{IEEE Robotics and Automation Letters}, 2023.

\bibitem{xie2018feedback}
Z.~Xie, G.~Berseth, P.~Clary, J.~Hurst, and M.~Van~de Panne, ``Feedback control for cassie with deep reinforcement learning,'' in \emph{2018 IEEE/RSJ International Conference on Intelligent Robots and Systems (IROS)}.\hskip 1em plus 0.5em minus 0.4em\relax IEEE, 2018, pp. 1241--1246.

\bibitem{xie2019iterative}
Z.~Xie, P.~Clary, J.~Dao, P.~Morais, J.~Hurst, and M.~van~de Panne, ``Iterative reinforcement learning based design of dynamic locomotion skills for cassie,'' \emph{arXiv preprint arXiv:1903.09537}, 2019.

\bibitem{kaup2024review}
M.~Kaup, C.~Wolff, H.~Hwang, J.~Mayer, and E.~Bruni, ``A review of nine physics engines for reinforcement learning research,'' \emph{arXiv preprint arXiv:2407.08590}, 2024.

\bibitem{fey2025bridging}
N.~Fey, G.~B. Margolis, M.~Peticco, and P.~Agrawal, ``Bridging the sim-to-real gap for athletic loco-manipulation,'' \emph{arXiv preprint arXiv:2502.10894}, 2025.

\bibitem{he2025asap}
T.~He, J.~Gao, W.~Xiao, Y.~Zhang, Z.~Wang, J.~Wang, Z.~Luo, G.~He, N.~Sobanbab, C.~Pan \emph{et~al.}, ``Asap: Aligning simulation and real-world physics for learning agile humanoid whole-body skills,'' \emph{arXiv preprint arXiv:2502.01143}, 2025.

\bibitem{aeran2016time}
A.~Aeran and H.~G. Lemu, ``Time integration schemes in dynamic problems-effect of damping on numerical stability and accuracy,'' in \emph{6th International Workshop of Advanced Manufacturing and Automation}.\hskip 1em plus 0.5em minus 0.4em\relax Atlantis Press, 2016, pp. 213--220.

\bibitem{starkit_roki2}
\BIBentryALTinterwordspacing
{STARKIT}, ``Roki-2,'' \url{https://starkit.su/roki-2/}, 2025. [Online]. Available: \url{https://starkit.su/roki-2/}
\BIBentrySTDinterwordspacing

\bibitem{hansen2016cma}
N.~Hansen, ``The cma evolution strategy: A tutorial,'' \emph{arXiv preprint arXiv:1604.00772}, 2016.

\bibitem{high2025torque}
\BIBentryALTinterwordspacing
``High torque,'' 2025. [Online]. Available: \url{https://github.com/HighTorque-Robotics}
\BIBentrySTDinterwordspacing

\bibitem{starkit_servo_alum}
\BIBentryALTinterwordspacing
STARKIT, ``Aluminum servo motor,'' \url{https://starkit.su/servo-alum/}, 2025. [Online]. Available: \url{https://starkit.su/servo-alum/}
\BIBentrySTDinterwordspacing

\bibitem{stable-baselines3}
\BIBentryALTinterwordspacing
A.~Raffin, A.~Hill, A.~Gleave, A.~Kanervisto, M.~Ernestus, and N.~Dormann, ``Stable-baselines3: Reliable reinforcement learning implementations,'' \emph{Journal of Machine Learning Research}, vol.~22, no. 268, pp. 1--8, 2021. [Online]. Available: \url{http://jmlr.org/papers/v22/20-1364.html}
\BIBentrySTDinterwordspacing

\end{thebibliography}

\begin{IEEEbiography}[{\includegraphics[width=1in,height=1.25in,clip,keepaspectratio]{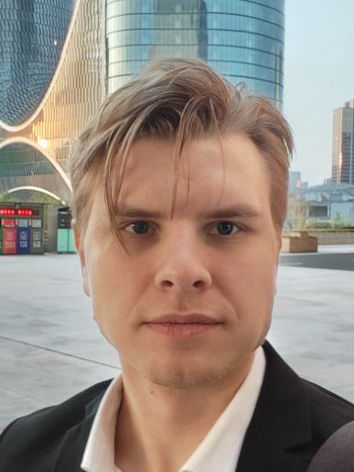}}]{Vyacheslav Kovalev}
is a researcher at the Moscow Institute of Physics and Technology (MIPT) and a Ph.D. student in Mathematics and Computer Science since 2024. He received the B.Sc. degree in Physics from Southern Federal University in 2021 and the M.Sc. degree in Mathematics and Computer Science from the Skolkovo Institute of Science and Technology in 2023. His research interests include actuator modeling and system identification, differentiable simulation and trajectory‑matching optimization, robot dynamics, and sim‑to‑real transfer for legged locomotion and control.
\end{IEEEbiography}

\begin{IEEEbiography}[{\includegraphics[width=1in,height=1.25in,clip,keepaspectratio]{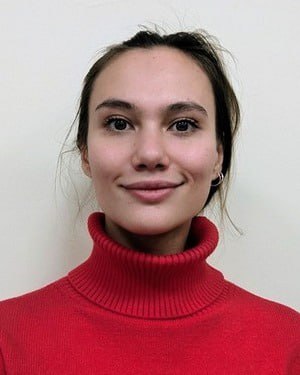}}]{Ekaterina Chaikovskaia} 
is a researcher at the Moscow Institute of Physics and Technology (MIPT). She received her B.Sc. degree in Automation of Technological Processes and Production from Bauman Moscow State Technical University in 2020, and her M.Sc. degree in Engineering Systems with a specialization in Robotics from the Skolkovo Institute of Science and Technology in 2022. 
Her research interests include reinforcement learning for locomotion, legged robotics, and control systems. Ekaterina is the author of seven scientific publications in the field of control systems.
\end{IEEEbiography}

\begin{IEEEbiography}[{\includegraphics[width=1in,height=1.25in,clip,keepaspectratio]{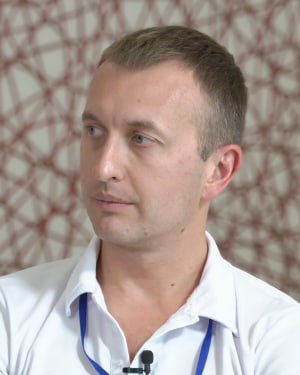}}]{Egor Davydenko}
is a researcher at the Moscow Institute of Physics and Technology (MIPT). He received his Ph.D. from Vladimir State University in 2009 and his M.Sc. from Yaroslavl State University in 2005. His research focuses on robotics, mechatronics, engineering, simulation, and computer vision, with particular emphasis on optimal control and reinforcement learning for legged robots.
His work aims to enhance robotic locomotion and perception in dynamic environments by integrating advanced control strategies, computer vision, and engineering principles. Egor is the author of 15 scientific publications and holds one invention. As part of the Starkit team, he contributed to winning first place in the RoboCup Worldwide Humanoid League in 2021, along with several other awards.
\end{IEEEbiography}

\begin{IEEEbiography}[{\includegraphics[width=1in,height=1.25in,clip,keepaspectratio]{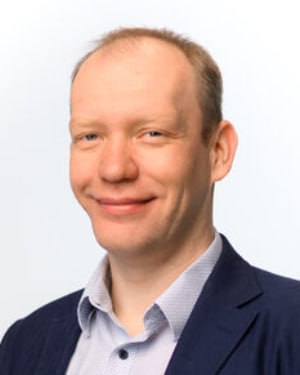}}]{Roman Gorbachev} 
is the Head of the Laboratory of Wave Processes and Control Systems and Director of the Research Center for Applied Artificial Intelligence Systems at the Moscow Institute of Physics and Technology (MIPT). He received his Ph.D. (Candidate of Technical Sciences) from MIPT in 2018, and his Bachelor's (2004) and Master's degrees (2008) from the Faculty of Radio Engineering and Cybernetics at MIPT.
His research interests include robotics, systems analysis, multi-agent system optimization, and the development of intelligent control systems, with a focus on collaborative manipulators and humanoid robotics. His work aims to advance adaptive control systems and integrate artificial intelligence methods into robotic platforms.
Roman is the author of over 20 scientific publications, holds 5 patents for inventions, 5 utility model patents, and 12 software copyrights. He also coordinates the Starkit team, which won the RoboCup World Championship in 2021 and is a multiple-time winner of regional championships.
\end{IEEEbiography}

\EOD

\end{document}